\batchmode
\makeatletter
\def\input@path{{/home/fyzhu/link2dropbox/self_Folder/myWorksOnDropboxs/201409_RRLbS//}}
\makeatother
\documentclass[twoside,twocolumn,english,10pt,twocolumn,twoside]{IEEEtran}
\usepackage[T1]{fontenc}
\usepackage[utf8]{inputenc}
\usepackage{color}
\definecolor{page_backgroundcolor}{rgb}{1, 1, 1}
\pagecolor{page_backgroundcolor}
\definecolor{document_fontcolor}{rgb}{0, 0, 0}
\color{document_fontcolor}
\usepackage{babel}
\usepackage{array}
\usepackage{float}
\usepackage{url}
\usepackage{multirow}
\usepackage{amsmath}
\usepackage{amsthm}
\usepackage{amssymb}
\usepackage{graphicx}
\usepackage[unicode=true,
 bookmarks=true,bookmarksnumbered=true,bookmarksopen=true,bookmarksopenlevel=1,
 breaklinks=false,pdfborder={0 0 0},backref=false,colorlinks=true]
 {hyperref}
\hypersetup{
 pdfauthor={Fei-yun Zhu}}

\makeatletter

\providecommand{\tabularnewline}{\\}
\floatstyle{ruled}
\newfloat{algorithm}{tbp}{loa}
\providecommand{\algorithmname}{Algorithm}
\floatname{algorithm}{\protect\algorithmname}

 \let\oldforeign@language\foreign@language
 \DeclareRobustCommand{\foreign@language}[1]{%
   \lowercase{\oldforeign@language{#1}}}
\theoremstyle{plain}
\newtheorem{thm}{\protect\theoremname}
\theoremstyle{plain}
\newtheorem{lem}[thm]{\protect\lemmaname}

\ifCLASSOPTIONcompsoc
\usepackage[caption=false,font=normalsize,labelfont=sf,textfont=sf]{subfig}
\else
\usepackage[caption=false,font=footnotesize]{subfig}
\fi

\usepackage{breakurl}
\usepackage{times}
\usepackage{epsfig}
\usepackage{multirow}
\usepackage{mdwlist}
\usepackage{algorithm}
\usepackage{algorithmic}% 
\usepackage{amsmath}
\usepackage{graphicx} 
\usepackage{url}
\usepackage[numbers,sort&compress]{natbib}

\@ifundefined{showcaptionsetup}{}{%
 \PassOptionsToPackage{caption=false}{subfig}}
\usepackage{subfig}
\makeatother

\providecommand{\lemmaname}{Lemma}
\providecommand{\theoremname}{Theorem}

\begin{document}
\global\long\def\mtbfA{\mathbf{A}}
 \global\long\def\mtbfa{\mathbf{a}}
 \global\long\def\mebfA{\bar{\mtbfA}}
 \global\long\def\mebfa{\bar{\mtbfa}}

\global\long\def\mhbfA{\widehat{\mathbf{A}}}
 \global\long\def\mhbfa{\widehat{\mathbf{a}}}
 \global\long\def\mtcalA{\mathcal{A}}

\global\long\def\mtbfB{\mathbf{B}}
 \global\long\def\mtbfb{\mathbf{b}}
 \global\long\def\mebfB{\bar{\mtbfB}}
 \global\long\def\mebfb{\bar{\mtbfb}}

\global\long\def\mhbfB{\widehat{\mathbf{B}}}
 \global\long\def\mhbfb{\widehat{\mathbf{b}}}
 \global\long\def\mtcalB{\mathcal{B}}

\global\long\def\mtbfC{\mathbf{C}}
 \global\long\def\mtbfc{\mathbf{c}}
 \global\long\def\mebfC{\bar{\mtbfC}}
 \global\long\def\mebfc{\bar{\mtbfc}}

\global\long\def\mhbfC{\widehat{\mathbf{C}}}
 \global\long\def\mhbfc{\widehat{\mathbf{c}}}
 \global\long\def\mtcalC{\mathcal{C}}
 \global\long\def\mtbbC{\mathbb{C}}

\global\long\def\mtbfD{\mathbf{D}}
 \global\long\def\mtbfd{\mathbf{d}}
 \global\long\def\mebfD{\bar{\mtbfD}}
 \global\long\def\mebfd{\bar{\mtbfd}}

\global\long\def\mhbfD{\widehat{\mathbf{D}}}
 \global\long\def\mhbfd{\widehat{\mathbf{d}}}
 \global\long\def\mtcalD{\mathcal{D}}

\global\long\def\mtbfE{\mathbf{E}}
 \global\long\def\mtbfe{\mathbf{e}}
 \global\long\def\mebfE{\bar{\mtbfE}}
 \global\long\def\mebfe{\bar{\mtbfe}}

\global\long\def\mhbfE{\widehat{\mathbf{E}}}
 \global\long\def\mhbfe{\widehat{\mathbf{e}}}
 \global\long\def\mtcalE{\mathcal{E}}
 \global\long\def\mtexpect{\mathbb{E}}

\global\long\def\mtbfF{\mathbf{F}}
 \global\long\def\mtbff{\mathbf{f}}
 \global\long\def\mebfF{\bar{\mathbf{F}}}
 \global\long\def\mebff{\bar{\mathbf{f}}}

\global\long\def\mhbfF{\widehat{\mathbf{F}}}
 \global\long\def\mhbff{\widehat{\mathbf{f}}}
 \global\long\def\mtcalF{\mathcal{F}}

\global\long\def\mtbfG{\mathbf{G}}
 \global\long\def\mtbfg{\mathbf{g}}
 \global\long\def\mebfG{\bar{\mathbf{G}}}
 \global\long\def\mebfg{\bar{\mathbf{g}}}

\global\long\def\mhbfG{\widehat{\mathbf{G}}}
 \global\long\def\mhbfg{\widehat{\mathbf{g}}}
 \global\long\def\mtcalG{\mathcal{G}}

\global\long\def\mtbfH{\mathbf{H}}
 \global\long\def\mtbfh{\mathbf{h}}
 \global\long\def\mebfH{\bar{\mathbf{H}}}
 \global\long\def\mebfh{\bar{\mathbf{h}}}

\global\long\def\mhbfH{\widehat{\mathbf{H}}}
 \global\long\def\mhbfh{\widehat{\mathbf{h}}}
 \global\long\def\mtcalH{\mathcal{H}}

\global\long\def\mtbfI{\mathbf{I}}
 \global\long\def\mtbfi{\mathbf{i}}
 \global\long\def\mebfI{\bar{\mathbf{I}}}
 \global\long\def\mebfi{\bar{\mathbf{i}}}

\global\long\def\mhbfI{\widehat{\mathbf{I}}}
 \global\long\def\mhbfi{\widehat{\mathbf{i}}}
 \global\long\def\mtcalI{\mathcal{I}}

\global\long\def\mtbfJ{\mathbf{J}}
 \global\long\def\mtbfj{\mathbf{j}}
 \global\long\def\mebfJ{\bar{\mathbf{J}}}
 \global\long\def\mebfj{\bar{\mathbf{j}}}

\global\long\def\mhbfJ{\widehat{\mathbf{J}}}
 \global\long\def\mhbfj{\widehat{\mathbf{j}}}
 \global\long\def\mtcalJ{\mathcal{J}}

\global\long\def\mtbfK{\mathbf{K}}
 \global\long\def\mtbfk{\mathbf{k}}
 \global\long\def\mebfK{\bar{\mathbf{K}}}
 \global\long\def\mebfk{\bar{\mathbf{k}}}

\global\long\def\mhbfK{\widehat{\mathbf{K}}}
 \global\long\def\mhbfk{\widehat{\mathbf{k}}}
 \global\long\def\mtcalK{\mathcal{K}}

\global\long\def\mtbfL{\mathbf{L}}
 \global\long\def\mtbfl{\mathbf{l}}
 \global\long\def\mebfL{\bar{\mathbf{L}}}
 \global\long\def\mebfl{\bar{\mathbf{l}}}

\global\long\def\mhbfL{\widehat{\mathbf{K}}}
 \global\long\def\mhbfl{\widehat{\mathbf{k}}}
 \global\long\def\mtcalL{\mathcal{L}}

\global\long\def\mtbfM{\mathbf{M}}
 \global\long\def\mtbfm{\mathbf{m}}
 \global\long\def\mebfM{\bar{\mathbf{M}}}
 \global\long\def\mebfm{\bar{\mathbf{m}}}

\global\long\def\mhbfM{\widehat{\mathbf{M}}}
 \global\long\def\mhbfm{\widehat{\mathbf{m}}}
 \global\long\def\mtcalM{\mathcal{M}}

\global\long\def\mtbfN{\mathbf{N}}
 \global\long\def\mtbfn{\mathbf{n}}
 \global\long\def\mebfN{\bar{\mathbf{N}}}
 \global\long\def\mebfn{\bar{\mathbf{n}}}

\global\long\def\mhbfN{\widehat{\mathbf{N}}}
 \global\long\def\mhbfn{\widehat{\mathbf{n}}}
 \global\long\def\mtcalN{\mathcal{N}}

\global\long\def\mtbfO{\mathbf{O}}
 \global\long\def\mtbfo{\mathbf{o}}
 \global\long\def\mebfO{\bar{\mathbf{O}}}
 \global\long\def\mebfo{\bar{\mathbf{o}}}

\global\long\def\mhbfO{\widehat{\mathbf{O}}}
 \global\long\def\mhbfo{\widehat{\mathbf{o}}}
 \global\long\def\mtcalO{\mathcal{O}}

\global\long\def\mtbfP{\mathbf{P}}
 \global\long\def\mtbfp{\mathbf{p}}
 \global\long\def\mebfP{\bar{\mathbf{P}}}
 \global\long\def\mebfp{\bar{\mathbf{p}}}

\global\long\def\mhbfP{\widehat{\mathbf{P}}}
 \global\long\def\mhbfp{\widehat{\mathbf{p}}}
 \global\long\def\mtcalP{\mathcal{P}}

\global\long\def\mtbfQ{\mathbf{Q}}
 \global\long\def\mtbfq{\mathbf{q}}
 \global\long\def\mebfQ{\bar{\mathbf{Q}}}
 \global\long\def\mebfq{\bar{\mathbf{q}}}

\global\long\def\mhbfQ{\widehat{\mathbf{Q}}}
 \global\long\def\mhbfq{\widehat{\mathbf{q}}}
\global\long\def\mtcalQ{\mathcal{Q}}

\global\long\def\mtbfR{\mathbf{R}}
 \global\long\def\mtbfr{\mathbf{r}}
 \global\long\def\mebfR{\bar{\mathbf{R}}}
 \global\long\def\mebfr{\bar{\mathbf{r}}}

\global\long\def\mhbfR{\widehat{\mathbf{R}}}
 \global\long\def\mhbfr{\widehat{\mathbf{r}}}
\global\long\def\mtcalR{\mathcal{R}}
 \global\long\def\mtbbR{\mathbb{R}}

\global\long\def\mtbfS{\mathbf{S}}
 \global\long\def\mtbfs{\mathbf{s}}
 \global\long\def\mebfS{\bar{\mathbf{S}}}
 \global\long\def\mebfs{\bar{\mathbf{s}}}

\global\long\def\mhbfS{\widehat{\mathbf{S}}}
 \global\long\def\mhbfs{\widehat{\mathbf{s}}}
\global\long\def\mtcalS{\mathcal{S}}

\global\long\def\mtbfT{\mathbf{T}}
 \global\long\def\mtbft{\mathbf{t}}
 \global\long\def\mebfT{\bar{\mathbf{T}}}
 \global\long\def\mebft{\bar{\mathbf{t}}}

\global\long\def\mhbfT{\widehat{\mathbf{T}}}
 \global\long\def\mhbft{\widehat{\mathbf{t}}}
 \global\long\def\mtcalT{\mathcal{T}}

\global\long\def\mtbfU{\mathbf{U}}
 \global\long\def\mtbfu{\mathbf{u}}
 \global\long\def\mebfU{\bar{\mathbf{U}}}
 \global\long\def\mebfu{\bar{\mathbf{u}}}

\global\long\def\mhbfU{\widehat{\mathbf{U}}}
 \global\long\def\mhbfu{\widehat{\mathbf{u}}}
 \global\long\def\mtcalU{\mathcal{U}}

\global\long\def\mtbfV{\mathbf{V}}
 \global\long\def\mtbfv{\mathbf{v}}
 \global\long\def\mebfV{\bar{\mathbf{V}}}
 \global\long\def\mebfv{\bar{\mathbf{v}}}

\global\long\def\mhbfV{\widehat{\mathbf{V}}}
 \global\long\def\mhbfv{\widehat{\mathbf{v}}}
\global\long\def\mtcalV{\mathcal{V}}

\global\long\def\mtbfW{\mathbf{W}}
 \global\long\def\mtbfw{\mathbf{w}}
 \global\long\def\mebfW{\bar{\mathbf{W}}}
 \global\long\def\mebfw{\bar{\mathbf{w}}}

\global\long\def\mhbfW{\widehat{\mathbf{W}}}
 \global\long\def\mhbfw{\widehat{\mathbf{w}}}
 \global\long\def\mtcalW{\mathcal{W}}

\global\long\def\mtbfX{\mathbf{X}}
 \global\long\def\mtbfx{\mathbf{x}}
 \global\long\def\mebfX{\bar{\mtbfX}}
 \global\long\def\mebfx{\bar{\mtbfx}}

\global\long\def\mhbfX{\widehat{\mathbf{X}}}
 \global\long\def\mhbfx{\widehat{\mathbf{x}}}
 \global\long\def\mtcalX{\mathcal{X}}

\global\long\def\mtbfY{\mathbf{Y}}
 \global\long\def\mtbfy{\mathbf{y}}
\global\long\def\mebfY{\bar{\mathbf{Y}}}
 \global\long\def\mebfy{\bar{\mathbf{y}}}

\global\long\def\mhbfY{\widehat{\mathbf{Y}}}
 \global\long\def\mhbfy{\widehat{\mathbf{y}}}
 \global\long\def\mtcalY{\mathcal{Y}}

\global\long\def\mtbfZ{\mathbf{Z}}
 \global\long\def\mtbfz{\mathbf{z}}
 \global\long\def\mebfZ{\bar{\mathbf{Z}}}
 \global\long\def\mebfz{\bar{\mathbf{z}}}

\global\long\def\mhbfZ{\widehat{\mathbf{Z}}}
 \global\long\def\mhbfz{\widehat{\mathbf{z}}}
\global\long\def\mtcalZ{\mathcal{Z}}

\global\long\def\mtvar{\mathbf{\text{Var}}}

\global\long\def\mtth{\text{th}}

\global\long\def\mtbfzero{\mathbf{0}}
 \global\long\def\mtbfone{\mathbf{1}}

\global\long\def\mttrace{\text{Tr}}

\global\long\def\mttotalVariation{\text{TV}}

\global\long\def\mtdet{\text{Det}}

\global\long\def\mtvec{\mathbf{\text{vec}}}

\global\long\def\mtvar{\mathbf{\text{var}}}

\global\long\def\mtcov{\mathbf{\text{cov}}}

\global\long\def\mtsubTo{\mathbf{\text{s.t.}}}

\global\long\def\mtfor{\text{for}}

\global\long\def\mtrank{\text{rank}}

\global\long\def\mtdiag{\mathbf{\text{diag}}}

\global\long\def\mtsign{\mathbf{\text{sign}}}

\global\long\def\mtloss{\mathbf{\text{loss}}}

\global\long\def\mtwhen{\text{when}}

\global\long\def\mtexpect{\mathbb{E}}

\global\long\def\mtcalN{\mathcal{N}}

\global\long\def\mtbbR{\mathbb{R}}

\title{Effective Spectral Unmixing via Robust Representation and Learning-based
Sparsity }

\author{Feiyun~Zhu, Ying~Wang, Bin~Fan, Gaofeng Meng and~Chunhong~Pan\thanks{Feiyun~Zhu, Ying~Wang, Bin~Fan, Gaofeng Meng and~Chunhong~Pan
are with the National Laboratory of Pattern Recognition, Institute
of Automation, Chinese Academy of Sciences (e-mail: fyzhu0915@gmail.com
and \{ywang, bfan, gfmeng and chpan\}@nlpr.ia.ac.cn).}}

\markboth{\#\#\#}{F. Y. Zhu \MakeLowercase{\emph{et al.}}: Spectral Unmixing via
RRLbS }
\maketitle
\begin{abstract}
Hyperspectral unmixing (HU) plays a fundamental role in a wide range
of hyperspectral applications. It is still challenging due to the
common presence of outlier channels and the large solution space.
To address the above two issues, we propose a novel model by emphasizing
both robust representation and learning-based sparsity. Specifically,
we apply the $\ell_{2,1}$-norm to measure the representation error,
preventing outlier channels from dominating our objective. As a result,
the side effects of outlier channels are largely relieved. Besides,
we observe that the mixed level of each pixel varies over image grids.
Based on this observation, we exploit a learning-based sparsity method
to simultaneously learn the HU results and a sparse guidance map.
Via this guidance map, the sparsity constraint in the $\ell_{p}\left(0<p\leq1\right)$-norm
is adaptively imposed according to the mixed level of each pixel.
Compared with state-of-the-art methods, our model is better suited
to the real situation, thus expected to achieve better HU results.
The resulted objective is highly non-convex and non-smooth, and so
it is hard to optimize. As a profound theoretical contribution, we
propose an efficient algorithm to solve it. Meanwhile, the convergence
proof and the computational complexity analysis are systematically
provided. Extensive evaluations verify that our method is highly promising
for the HU task---it achieves very accurate guidance maps and much
better HU results compared with state-of-the-art methods. \end{abstract}


\section{Introduction}

\IEEEPARstart{H}{yperspectral} unmixing (HU) is one of the most
foundation steps for various applications, such as sub-pixel mapping~\cite{Mertens_2003_IJRS_subPixelMapping},
high-resolution hyperspectral imaging~\cite{Kawakami_2011_CVPR_highResolutionHyperImaging},
hyperspectral enhancement~\cite{zhGuo_2009_SPIE_L1unmixing&Enhancement},
hyperspectral compression and reconstruction~\cite{cbLi_2012_Tip_CompressSensing&Unmixing},
hyperspectral visualization and understanding~\cite{ShangshuCai_2007_TGRS_hyperVisualizationUsingDoubleLayers,fyzhu_2014_TIP_DgS_NMF},
detection and identification substances in the scene~\cite{zhGuo_2009_SPIE_L1unmixing&Enhancement,Qian_11_TGRS_NMF+l1/2}
etc.  The goal  of HU is to break down each pixel spectrum into
a set of ``pure'' spectra (called \emph{endmembers} such as the spectra
of water, grass, tree etc.), weighted by the corresponding proportions,
called \emph{abundances}. Formally, HU methods take in a hyperspectral
image with $L$ channels and $N$ pixels\ \cite{haichangLi_2016_IJRS_LablePropagationHyperClassification},
and assume that each pixel $\mtbfx\in\mtbbR_{+}^{L}$ is a composite
of $K$ \emph{endmembers} $\left\{ \mtbfm_{k}\right\} _{k=1}^{K}\in\mtbbR_{+}^{L}$.
Specifically, the linear combinatorial model is the most popular one
\begin{equation}
\mtbfx=\sum_{k=1}^{K}\mtbfm_{k}a_{k},\quad\mtsubTo\:a_{k}\geq0\ \text{and\ }\sum_{k=1}^{K}a_{k}=1,\label{eq:LinearMixedModel_vector}
\end{equation}
where $a_{k}$ is the composite \emph{abundance} of the $k^{\mtth}$
\emph{endmember}. In the unsupervised setting, both \emph{endmembers}
$\left\{ \mtbfm_{k}\right\} _{k=1}^{K}$ and \emph{abundances} $\left\{ a_{k}\right\} _{k=1}^{K}$
are unknown. Such case makes the objective function non-convex and
the solution space very large~\cite{fyzhu_2014_TIP_DgS_NMF,LiuXueSong_2011_TGRS_ConstrainedNMF}.
Therefore, reasonable prior knowledge is required to restrict the
solution space, and moreover to bias the solution toward good stationary
points.
\begin{figure}[t]
\centering{}\includegraphics[width=0.98\columnwidth]{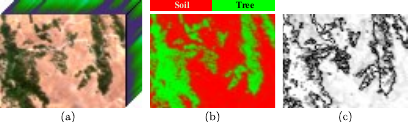}\caption{As the mixed level of each pixel varies over image grids, the sparse
constraint should be individually imposed according to the mixed level
of each pixel, rather than roughly imposed at the same strength for
all the pixels. (a) hyperspectral image of two substances: soil and
tree. (b) \emph{abundance} map. (c) is the guided map exhibiting the
individually mixed level of each pixel---the darker indicates the
more mixed. (best viewed in color)\label{fig:illustrate_individual_mixedLevel}}
\end{figure}

To reduce the solution space, various constraints are imposed upon
the\emph{ abundance}~\cite{LiuXueSong_2011_TGRS_ConstrainedNMF,Qian_11_TGRS_NMF+l1/2,Jia_09_TGRS_ConstainedNMF,JmLiu_12_SlectedTopics_W-NMF,Cai_11_PAMI_GNMF}
as well as upon the \emph{endmember}~\cite{Miao_07_ITGRS_NMFMVC,nWang_13_SelectedTopics_EDC-NMF}.
Although, these methods work to some extent, they are far from the
optimal for the following two reasons. First, the side effects of
badly degraded channels are generally ignored in state-of-the-art
methods. Many degraded channels deviate significantly from the majority
of hyperspectral channels. The objective of state-of-the-art methods
is easily dominated by the outlier channel, leading to poor unmixing
performances. Second, almost all existing constraints are roughly
imposed at the same strength for all the factors. Such implementation
does not meet the practical situation; an example is illustrated in
Fig.\ \ref{fig:illustrate_individual_mixedLevel}, where the mixed
level of each pixel varies over image grids. Therefore, it is more
reasonable to impose an individual strength of sparse constraints
according to the mixed level of each pixel, rather than roughly impose
the same strength of sparse constraints for all the pixel. Please
refer to the footnote\footnote{Note that the mixed level of a pixel $\mtbfx_{n}$ is negatively correlated
with the sparse level of the corresponding \emph{abundance $\mtbfa_{n}$.}
If pixel $\mtbfx_{n}$ is very ``pure'', the \emph{abundance $\mtbfa_{n}$
}will be highly sparse and vice versa. For example in Fig.~\ref{fig:illustrate_individual_mixedLevel},
the \emph{abundances} are $\left[0.5,0.5\right]$ and $\left[0,9,0.1\right]$
respectively for pixel $\mtbfy_{i}$ and $\mtbfy_{j}$. Then $\mtbfy_{i}$
is more mixed and less sparser than $\mtbfy_{j}$. } for the detailed explanation of the relationship between mixing pixel
and sparsity.

Indeed, there is one method\ \cite{fyzhu_2014_TIP_DgS_NMF} imposing
the adpatively sparse constraint for each pixel. However, \cite{fyzhu_2014_TIP_DgS_NMF}
proposed a heuristic method to learn the guided map which is ineffective
and inaccurate for the vast smooth areas in the image. Accordingly,
the sparse constraint is inaccurate. It is expected that the more
accurate constraints would bias the solution to the more satisfactory
local minima.

To alleviate the above issues, we propose a novel method, named effective
spectral unmixing via robust representation and learning-based sparsity
(RRLbS) for the HU task. Specifically, the $\ell_{2,1}$-norm is employed
to measure the representation loss, preventing large errors from dominating
our objective. In this way, the robustness against outlier channels
is greatly enhanced. Besides, a learning-based sparsity method is
exploited to individually impose the sparsity constraint according
to the mixed level of each pixel. The main contributions of this work
are summarized as follows.
\begin{itemize}
\item It is the side influences of badly degraded channels that are generally
ignored in the state-of-the-art methods. To the best of our knowledge,
this is the first attempt in the HU field to propose the $\ell_{2,1}$-norm
based robust model to relieve the side effects of outlier channels. 
\item Besides, we propose a novel learning-based sparsity method to simultaneously
learn the HU results and a guided map. The method to esimate the guided
map is novel and effective. Through this guided map, the mixed level
of every pixel is described and respected by imposing an adaptive
sparsity constraint according to the mixed level of each pixel. Such
implementation helps to achieve highly promising HU results and guided
maps. 
\item We propose an efficient algorithm to solve the joint $\ell_{2,1}$-norm
and $\ell_{p}$-norm based objective, which is highly non-convex,
non-smooth and challenging to solve. Both theoretical and empirical
analyses are conducted to verify its convergence property. Besides,
the computational complexity analysis is systematically analyzed as
well. 
\end{itemize}
The rest of this paper is organized as below: in Section~\ref{sec:PreviousWork},
the related HU work is systematically reviewed. Section~\ref{sec:modeling_for_RRLbS}
presents the new model (RRLbS) and its physical motivations. The algorithm
as well its theoretical convergence proof and computational complexity
analysis are given in Section~\ref{sec:algorithm_for_RRLbS}. Then,
extensive evaluations are provided in Section~\ref{sec:Evaluation}.
Finally, the conclusion of this work is drawn in Section~\ref{sec:Conclusions}.

\section{\label{sec:PreviousWork}Previous Work}

The existing HU methods are typically categorized into three types:
supervised methods\,\cite{zhGuo_2009_SPIE_L1unmixing&Enhancement,cbLi_2012_Tip_CompressSensing&Unmixing},
weakly supervised methods\,\cite{Iordache_2011_TGRS_SparseUnmixing,Iordache_2012_TGRS_TV_SparseUnmixing}
and unsupervised methods\,\cite{Bayliss_1997_SPIE_ICA_unmixing,SJia_07_TGRS_SSCBSS,Jia_09_TGRS_ConstainedNMF,Qian_11_TGRS_NMF+l1/2,nWang_13_SelectedTopics_EDC-NMF,fyzhu_2014_IJPRS_SSNMF}
(Here the defintions of supervised, weakly supervised are very different
from the definition in machine learning\ \cite{guangliangCheng_2014_ICIP,yaoyao_2017_MICCAI,guangliangCheng_2015_ICIP,xiaoping_2017_ICASSP,zhengxu_2017_ACMBCB,guangliangCheng_2016_neurocomputing,xinliang_2017_CVPR_WSISA,guangliangCheng_2016_TGRSL,fyzhu_2017_RLDM_WarmStart}.
Please refer to the following paragraphs for their new definitions).
For the supervised methods, the \emph{endmembers} are given in advance;
only the \emph{abundances} need to estimate. Although in this way,
the HU problem is greatly simplified, it is usually inconvenient or
intractable to obtain feasible \emph{endmembers} in the supervised
setting, thus, hampering the acquisition of good estimations. 

Accordingly, the weakly supervised methods\,\cite{Iordache_2011_TGRS_SparseUnmixing,Iordache_2012_TGRS_TV_SparseUnmixing}
are a type of popular methods. A large library of substance spectra
have been collected by a field spectrometer beforehand. Then, the
HU task becomes the problem of finding the optimal subset of spectra
in the library that can best represent all pixels in the scene\,\cite{Iordache_2011_TGRS_SparseUnmixing}.
Unfortunately, the library is far from optimal for the fact that the
spectra in it are not standardly unified. First, for different hyperspectral
sensors, the spectral shape of the same substance are greatly inconsistent.
Second, for various hyperspectral images, the length of pixel spectra
is largely different as well---for example some images have $115$
channels, while another has $224$ channels and some even have $480$
spectral channes. Their electromagnetic spectra ranges are also very
different. Finally, the recording conditions are highly different
as well---some hyperspectral images are captured far from the outer
space, while some hyperspectral images are obtained from the airplane
or ground. Due to the atmospheric effects etc., the different recording
condition would lead to different spectral appearances. In short,
the weakness of the library brings side effects on this kind of methods. 

More commonly, the \emph{endmembers} are selected from the image itself
to ensure the spectral consistency~\cite{zhGuo_2009_SPIE_L1unmixing&Enhancement}
and the unsupervised HU methods are preferred. The unsupervised HU
methods could be generally  categorized into two types: geometric
methods~\cite{Boardman_1995_JPL_PPI,Jose_05_TGRS_Vca,Chang_2006_TGRS_SGA,junLi_2008_IGARSS_MVSA,Bioucas_2009_WHISPERS_SISAL,Martin_12_SelectedTopics_SSPP}
and statistical ones~\cite{Wang_06_ITGRS_ICAextEndmember,Dobigeon_ISP_BaysianHU,Jose_09_WHISPERS_SplittingAugmentedLag,Jose_12_TGRS_HUbyDirchletComponents,Chanussot_2014_TGRS_NonlinearUnmixing,fyzhu_2014_IJPRS_SSNMF,fyzhu_2014_TIP_DgS_NMF}.
The geometric methods usually exploit the simplex to model the distribution
of spectral pixels.  Perhaps, N-FINDR~\cite{Michael_99_PSCIS_nFindr}
and Vertex Component Analysis (VCA)~\cite{Jose_05_TGRS_Vca} are
the most typical geometric methods. For N-FINDR, the \emph{endmembers}
are extracted by inflating a simplex inside the hyperspectral pixel
distribution and treating the vertices of a simplex with the largest
volume as \emph{endmembers}~\cite{Michael_99_PSCIS_nFindr}. VCA~\cite{Jose_05_TGRS_Vca}
projects all pixels onto a direction orthogonal to the simplex spanned
by the chosen\emph{ endmembers}; the new \emph{endmember} is identified
as the extreme of the projection. Although these methods are simple
and fast, they suffer from the requirement of pure pixels, which
is usually unreliable in practice\,\cite{Jia_09_TGRS_ConstainedNMF,Qian_11_TGRS_NMF+l1/2,XLu_2013_TGRS_ManifoldSparseNMF}. 

Accordingly, many statistical methods have been proposed for or applied
to the HU problem, among which the Nonnegative Matrix Factorization
(NMF)~\cite{Lee_99_Nature_NMF} and its extensions are the most popular.
The goal of NMF is to find two nonnegative matrices to approximate
the original matrix with their product~\cite{Cai_11_PAMI_GNMF}.
There are two valuable reasons to apply the nonnegative constraint
on both factor matrices. First, both \emph{endmembers} and \emph{abundances
}should be nonnegative. Such case means that the NMF model is physically
suited to the HU task. Second, the nonnegative constraint only allows
for additive combinations, not subtractions, yielding a parts-based
representation\,\cite{Cai_11_PAMI_GNMF}. This parts-based property
enables factor results more intuitive and interpretable, as existing
studies on psychological and physiological field have shown human
brain also works in the parts-based manner~\cite{Palmer_77_CP_perceptualRepresent,Logothetis_96_ARN_VisObjRecognition}. 

Although NMF is well adapted to applications of face analysis~\cite{Stanzli_01_CVPR_locNMF,Roman_11_PAMI_EarthMoverNMF}
and documents clustering~\cite{WeiXu_03_SIGIR_DocClusterNMF,Farial_06_InfProManag_DocumNMF},
the objective function is non-convex, naturally resulting in a large
solution space~\cite{Lee_00_NIPS_NMF}. Many extensions have been
proposed by employing all kinds of priors to restrict the solution
space. For the HU task, the priors are imposed either upon \emph{abundances\,}\cite{Qian_11_TGRS_NMF+l1/2,JmLiu_12_SlectedTopics_W-NMF,LiuXueSong_2011_TGRS_ConstrainedNMF}
or upon\emph{ endmembers\,}\cite{Miao_07_ITGRS_NMFMVC,nWang_13_SelectedTopics_EDC-NMF}.
For example, the Local Neighborhood Weights regularized NMF method~\cite{JmLiu_12_SlectedTopics_W-NMF}
(W-NMF) assumes that the hyperspectral pixels are distributed on a
manifold structure and exploits appropriate weights in the local neighborhood
to enhance the spectral and spatial information\,\cite{Prasad_2014_SPM_ManifoldFeatureExtraction}.
This information could be eventually transferred to the \emph{abundance}
space via the Laplace graph constraint. Actually, this constraint
has a smooth impact. It will weaken the parts-based property of NMF. 

Inspired by MVC-NMF~\cite{Miao_07_ITGRS_NMFMVC}, Wan et al.~\cite{nWang_13_SelectedTopics_EDC-NMF}
proposed the EDC-NMF method. The basic assumption is originated from
two perspectives. First, due to the high spectral resolution of hyperspectral
sensors, the \emph{endmember} signal should be smooth itself. Besides,
the \emph{endmember} signals should possess distinct shapes so that
we can separate out different materials\ \cite{zhGuo_2009_SPIE_L1unmixing&Enhancement}.
However, in their algorithm, they take a derivative along the \emph{endmember}
vector, introducing negative values to the updating rule. To make
up this drawback, the elements in the \emph{endmember} matrix are
required to project to a given nonnegative value after each iteration.
Consequently, the regularization parameters could not be chosen freely,
which limits the efficacy of this method. 

The sparsity-based methods are the most successful methods for the
HU task. They assume that in hyperspectral images most pixels are
mixed by a subset of \emph{endmembers}, rather than all \emph{endmembers},\emph{
}thus employing various kinds of sparse constraints on \emph{abundances}.
Specifically, the $\ell_{1/2}$-NMF~\cite{Qian_11_TGRS_NMF+l1/2}
is a state of the art sparsity regularized NMF method that is derived
from Hoyer's lasso regularized NMF~\cite{Hoyer_02_NNSP_NMF_l1}.
The lasso constraint~\cite{Tibshirani_94_Statist_Lasso,Donoho_96_ITIT_CS}
could not enforce further sparse when the full additivity constraint
is used, limiting the effectiveness of this method~\cite{Qian_11_TGRS_NMF+l1/2}.
Thus, Qian et al. exploits the $\ell_{p}\left(p=1/2\right)$-norm
to regularize the \emph{abundances} as it has been proved by Fan et
al.~\cite{Fan_2001_JASA_LaNorm} that the $\ell_{p}\left(p=1/2\right)$
constraint could obtain sparser solutions than the $\ell_{1}$-norm
does.

Although the related methods work to some extent, they are far from
the optima. Thus, we propose a new method by emphasizing both robust
representation and learning-based sparsity. Through the former, the
side effects of outlier channels are greatly relieved. While via the
latter, it is more likely to bias the HU solution to some suited stationary
points in the large solution space.

\section{\label{sec:modeling_for_RRLbS}RRLbS: Robust Representation and Learning-based
Sparsity }

In this section, we propose a novel model by emphasizing both robust
representation and learning-based sparsity. To relieve the side influences
of badly degraded channels, the $\ell_{2,1}$-norm, rather than $\ell_{2}$-norm,
is employed to measure the representation\emph{ }loss, preventing
too large errors from dominating our objective. Then, a learning-based
sparsity method is proposed  to update a guidance map, by which the
sparse constraint could be individually imposed according to the mixed
level of each pixel. Such implementation is more reasonable, thus
expected to get better HU results.

\textbf{Notation.} In this paper, we use boldface uppercase letters
to denote matrices and boldface lowercase letters to represent vectors.
Given a matrix $\mtbfX\triangleq\left\{ X_{ln}\right\} \in\mtbbR^{L\times N}$,
we denote the $l^{\mtth}$ row and $n^{\mtth}$ column as $\mtbfx^{l}\in\mtbbR^{1\times N}$
and $\mtbfx_{n}\in\mtbbR^{L}$ respectively. $X_{ln}$ is the $\left(l,n\right)$-th
entry in the matrix. A nonnegative matrix is denoted as $\mtbfX\geq\mtbfzero$
or $\mtbfX\in\mtbbR_{+}^{L\times N}$. The $\ell_{2,1}$-norm of matrices
is defined as $\left\Vert \mtbfX\right\Vert _{2,1}\!=\!\sum_{l}^{L}\left(\sum_{n}^{N}X_{ln}^{2}\right)^{1/2}$. 

\textbf{Problem formalization. }In the HU problem, we are often given
a hyperspectral image of $N$ pixels and $L$ channels, which is denoted
by a nonnegative matrix $\mtbfX\triangleq\left[\mtbfx_{1},\mtbfx_{2},\cdots,\mtbfx_{N}\right]\in\mtbbR_{+}^{L\times N}$.
From the linear mixture perspective, the goal of HU is to find two
nonnegative matrices to well approximate $\mtbfX$ with their product.
Formally, the discrepancy between $\mtbfX$ and its representation
$\widetilde{\mtbfX}$ is modeled as 
\begin{equation}
\min_{\mtbfM,\mtbfA}\mtloss\left\{ \mtbfX,\widetilde{\mtbfX}\right\} ,\quad\mtsubTo\ \widetilde{\mtbfX}=\mtbfM\mtbfA,\mtbfM\geq\mtbfzero,\mtbfA\geq\mtbfzero,\label{eq:generalized_HU_loss}
\end{equation}
where $\mtbfM\triangleq\left[\mtbfm_{1},\cdots,\mtbfm_{K}\right]\in\mtbbR_{+}^{L\times K}$
is the \emph{endmember} matrix including $K$ spectral bases, $K\ll\min\left\{ L,N\right\} $;
$\mtbfA\triangleq\left[\mtbfa_{1},\cdots,\mtbfa_{N}\right]\in\mtbbR_{+}^{K\times N}$
is the corresponding\emph{ abundance} matrix---the $n^{\mtth}\left(\forall n\!=\!1,\cdots,N\right)$
column vector $\mtbfa_{n}$ contains all $K$ \emph{abundances} at
pixel $\mtbfx_{n}$; $\mtloss\left\{ \cdot,\cdot\right\} $ is a loss
function measuring the difference between two terms. When setting
$\mtloss\left\{ \cdot,\cdot\right\} $ as the Euclidean loss, the
objective\,\eqref{eq:generalized_HU_loss} becomes the standard NMF
problem\,\cite{fyzhu_2014_TIP_DgS_NMF,Lee_99_Nature_NMF,Lee_00_NIPS_NMF},
which is commonly used in a great number of state of the art HU methods\,\cite{LiuXueSong_2011_TGRS_ConstrainedNMF,Miao_07_ITGRS_NMFMVC,Jia_09_TGRS_ConstainedNMF,fyzhu_2014_TIP_DgS_NMF,Qian_11_TGRS_NMF+l1/2,nWang_13_SelectedTopics_EDC-NMF}.
However, the Euclidean loss is prone to outliers\,\cite{fpNie_2010_NIPS_JointL21_featureSelection,huaWang_2014_ICML_RobustMetricLearning}.
Accordingly, our initial goal is to propose a robust HU model. 
\begin{figure}[t]
\centering{}\includegraphics[width=0.7\columnwidth]{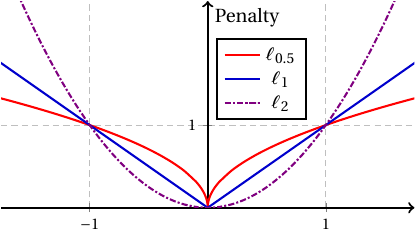}\caption{$\ell_{p}\left(0.5\leq p\leq1\right)$-norm versus $\ell_{2}$-norm
in shape. Compared with the $\ell_{2}$-norm, $\ell_{1}$ is more
capable of preventing large errors from dominating the objective energy.
Compared with the $\ell_{1}$ sparse constraint, $\ell_{p}\left(0.5\leq\!p\!<\!1\right)$
tends to find a sparser solution\ \cite{Fan_2001_JASA_LaNorm,Fan_2004_ANN-Statist_LaNorm,Qian_11_TGRS_NMF+l1/2}.
\label{fig:demo_Lp_loss}}
\end{figure}

\subsection{Robust Representation in the $\ell_{2,1}$-norm}

Existing HU methods generally use the Euclidean loss to measure the
representation error, that is
\begin{equation}
\min_{\mtbfM,\mtbfA}\sum_{l=1}^{N}\left\Vert \mtbfx^{l}-\mtbfm^{l}\mtbfA\right\Vert _{2}^{2},\quad\mtsubTo\ \mtbfM\geq\mtbfzero,\mtbfA\geq\mtbfzero,\label{eq:traditional_L2_loss_row}
\end{equation}
where $\mtbfx^{l}$ is the $l^{\mtth}$ channel (i.e. row vector)
in $\mtbfX$; $\mtbfm^{l}$ is the $l^{\mtth}$ channel in $\mtbfM$.
 Similar to the existing least square minimization based models in
machine learning and statistics\,\cite{FpNie_2013_IJCAI_EarlyActiveLearning,kaiYu_2006_ICML_activeLearning},
\eqref{eq:traditional_L2_loss_row} is sensitive to the presence of
outlier channels\,\cite{huaWang_2014_ICML_RobustMetricLearning,guangliangCheng_2016_JStars_robustHyperClassification,yingWang_2015_TIP_RobustUnmixing}.
 However, from the perspective of remote sensing, hyperspectral images
are very likely to contain outlier channels. This is  owing to the
following two reasons. First, due to the high spectral resolution
of hyperspectral sensors, it receives very litter energy from a narrow
wavelength range when producing each hyperspectral channel. In this
way, the imaging information is highly easy to be overwhelmed by
various kinds of noises. Second, the\textbf{ }bad imaging conditions
are responsible for the degraded channels as well---when imaging from
the outer space or airplanes, due to the water vapor and the atmospheric
effects etc., the hyperspectral channels are easy to be blank or badly
noised. Specifically, many noised channels deviate significantly from
the majority of the hyperspectral channels. They are actually outlier
channels.

To identify the outlier channels and to relieve the side effects they
cause, a robust loss in the $\ell_{2,1}$-norm is proposed
\begin{equation}
\min_{\mtbfM,\mtbfA}\sum_{l=1}^{L}\left\Vert \mtbfx^{l}-\mtbfm^{l}\mtbfA\right\Vert _{2},\quad\mtsubTo\ \mtbfM\geq\mtbfzero,\mtbfA\geq\mtbfzero.\label{eq:RRLbS_robust_L2p_loss_row}
\end{equation}
 Considering all the row vectors together, the objective\ \eqref{eq:RRLbS_robust_L2p_loss_row}
becomes the concise matrix format:
\begin{equation}
\min_{\mtbfM,\mtbfA}\left\Vert \mtbfX-\mtbfM\mtbfA\right\Vert _{2,1},\quad\mtsubTo\ \mtbfM\geq\mtbfzero,\mtbfA\geq\mtbfzero,\label{eq:RRLbS_robust_L2p_loss_matrix}
\end{equation}
In our new model, the $\ell_{2,1}$-norm is applied to the representation
loss---the $\ell_{1}$-norm is imposed among channels and the $\ell_{2}$-norm
is used for pixels. As the $\ell_{1}$-loss is capable of preventing
large representation errors to dominate the objective, as shown in
Fig.\,\ref{fig:demo_Lp_loss}. The side effects of outlier channels
are greatly reduced and the robustness of the HU task is enhanced.

\subsection{\label{sub:FineTuned-DgMap}Learning-based Sparsity Constraint via
the Guidance Map }

State-of-the-art methods generally impose an identical strength of
sparsity constraints for all the pixels, e.g., 
\[
\mtcalJ\left(\mtbfA\right)=\sum_{n=1}^{N}h_{n}\left\Vert \mtbfa_{n}\right\Vert _{1},\quad\mtcalJ\left(\mtbfA\right)=\sum_{n=1}^{N}h_{n}\left\Vert \mtbfa_{n}\right\Vert _{1/2}^{1/2},
\]
where $\left\{ h_{n}=1\right\} _{n=1}^{N}$ is the guided value for
both $\ell_{1}$-NMF\ \cite{Hoyer_02_NNSP_NMF_l1} and $\ell_{1/2}$-NMF\ \cite{Qian_11_TGRS_NMF+l1/2}.
However, the mixed level of each pixel varies over image grids, as
shown in Fig.\,\ref{fig:illustrate_individual_mixedLevel}. It is
more reasonable to impose an individual sparsity constraint according
the mixed level of each pixel.  To this end, we propose an iterative
method to learn a guidance map $\mtbfh\in\mtbbR_{+}^{N}$, by which
the sparsity constraint in the $\ell_{p}\left(0.5\leq p\leq1\right)$\footnote{Fan et al. has shown that the sparsity of $\ell_{p}\left(0.5\leq p\leq1\right)$
solution increases as $p$ decreases, whereas the sparsity of the
solution of $\ell_{p}\left(0<p\leq0.5\right)$ shows little change
with respect to $p$~\cite{Fan_2001_JASA_LaNorm,Qian_11_TGRS_NMF+l1/2}.
Thus, to ensure the sensitivity of the individually sparsity constraint,
it is sufficient to use the $\ell_{p}\left(0.5\!\leq\!p\!\leq\!1\right)$-norm
in our model. }-norm will be individually applied as 
\begin{equation}
\mtcalJ\left(\mtbfA\right)=\sum_{n=1}^{N}\left\Vert \mtbfa_{n}\right\Vert _{1-h_{n}}^{1-h_{n}}=\sum_{n=1}^{N}\sum_{k=1}^{K}\left|A_{kn}\right|^{1-H_{kn}},\label{eq:RRLbS_individual_sparseConstraint_vector}
\end{equation}
where $\left\Vert \mtbfa\right\Vert _{p}^{p}=\sum_{k}^{K}\left|a_{k}\right|^{p}$,
$h_{n}$ is the $n^{\mtth}$ entry in the guided map $\mtbfh$, reflecting
the mixed level of the $n^{\mtth}$ pixel; $H_{kn}$ is the $\left(k,n\right)$-th
element in the matrix $\mtbfH=\mtbfone_{K}\mtbfh^{T};$ $\mtbfone_{K}$
is the column vector of all ones with length $K$. For each pixel
$\mtbfx_{n}$, the choice of $p$ is solely dependent upon the corresponding
guidance value $h_{n}$, performing a nonlinearly weighting role for
the sparsity constraint upon $\mtbfa_{n}$. Specifically, as the sparsity
of $\ell_{p}$ solution increases as $p$ decreases\ \cite{Fan_2001_JASA_LaNorm,Fan_2004_ANN-Statist_LaNorm,Qian_11_TGRS_NMF+l1/2},
a smaller $1-h_{n}\left(\forall n=1,2,\cdots,N\right)$ will impose
a stronger sparse constraint on $\mtbfa_{n}$. In this way, the sparse
constraint is individually imposed for each pixel. The matrix format
of\ \eqref{eq:RRLbS_individual_sparseConstraint_vector} is
\begin{equation}
\mtcalJ\left(\mtbfA\right)=\left\Vert \mtbfA^{\mtbfone-\mtbfH}\right\Vert _{1},\label{eq:RRLbS_individual_sparseConstraint_matrix}
\end{equation}
where $\mtbfA^{\mtbfone-\mtbfH}\!=\!\left[\left(A_{kn}\right)^{1-H_{kn}}\right]\in\mtbbR_{+}^{K\times N}$
is an element-wise exponential operation. 

The remaining problem is how to learn the optimal guidance map $\mtbfh^{*}\in\mtbbR_{+}^{N}$.
In\ \cite{fyzhu_2014_TIP_DgS_NMF}, there is one heuristic method
to learn the guided map, which is effective in the transitional areas.
However, it is ineffective in the vast smooth areas in the image due
to the heuristic mechanism. An inaccurate guided map would lead to
an unsuitable sparse constraint. In this paper, we find that $\mtbfh^{*}$
is crucially dependent upon the mixed level of each pixel, i.e., the
sparse level of the optimal \emph{abundance} $\mtbfA^{*}$. In other
words, if pixel $\mtbfx_{n}$ is highly mixed (i.e. the \emph{abundance}
$\mtbfa_{n}^{*}$ is weakly sparse), $h_{n}$ will be small; once
$\mtbfx_{n}$ is highly ``pure'' (i.e. $\mtbfa_{n}^{*}$ is largely
sparse), $h_{n}$ will be large. However, the mixed level of each
pixel is unavailable due to the unknown of the optimal \emph{abundance}. 

To achieve a good guidance map, here we will propose a two-step strategy.
First, a heuristic strategy is used to get an initial guess. Then
a learning-based updating rule is exploited to generate a sequence
of improved estimates until they reach a stable solution. We observe
that pixels in the transition area or image edges are very likely
be highly mixed\,\cite{fyzhu_2014_TIP_DgS_NMF}. Accordingly, we
propose a heuristic strategy to get an initial guidance map:
\begin{equation}
h_{i}=\sum_{j\in\mtcalN_{i}}s_{ij},\quad\forall i\in\left\{ 1,2,\cdots N\right\} ,\label{eq:initialDgMap}
\end{equation}
where $\mtcalN_{i}$ is the neighborhood of $\mtbfx_{i}$ that includes
four neighbors; $s_{ij}$ is a similarities measured as
\[
s_{ij}=\exp\left(-\frac{\left\Vert \mtbfx_{j}-\mtbfx_{i}\right\Vert _{2}^{2}}{\sigma}\right),
\]
 $\sigma\in\left[0.005,0.08\right]$ is an easily tuned parameter.
In this way, the pixels in the transition area are treated as mixed
pixels. However, there is one vital problem with\ \eqref{eq:initialDgMap}---this
heuristic strategy could only tackle with pixels in the sudden change
area. For the vast smooth areas, the mixed information is intractable
to achieve. Thus, other strategies are required. 

Perhaps, the most direct clue to obtain the guided maps is the intrinsical
correlation with the optimal\emph{ abundance $\mtbfA^{*}$}, that
is $h_{n}^{*}=S\left(\mtbfa_{n}^{*}\right),\forall n\in\left\{ 1,2,\cdots,N\right\} $,
where 
\[
S:\left(\bigcup_{K\ge1}\mtbbR_{+}^{K}\right)\longrightarrow\mtbbR_{+}
\]
 is a sparsity measure that maps real vectors to a nonnegative value\ \cite{Hurley_2009_TIT_ComparingSparseMeasures}.
Although the optimal \emph{abundance} $\mtbfA^{*}$ is unavailable,
the updated \emph{abundance $\mtbfA^{\left(t\right)}$} is available
after the $t^{\mtth}$ iteration. If $\mtbfA^{\left(t\right)}\rightarrow\mtbfA^{*}$
over iteration steps, we could always generate a sequence of improved
estimates towards the optimal guidance map, that is $\mtbfh^{\left(t\right)}\rightarrow\mtbfh^{*}$,
by using the dependence of $h_{n}^{\left(t\right)}=S\left(\mtbfa_{n}^{\left(t\right)}\right),\mtbfh^{\left(t\right)}=\left\{ h_{n}^{\left(t\right)}\right\} _{n=1}^{N}$.
In turn, the learnt $\mtbfh^{\left(t\right)}$ helps to impose an
improved individual sparsity constraint for each pixel, eventually
leading to a more reliable unmixing result. This iterative process
is supposed to generate a sequence of ever improved estimates until
convergences.

Specifically, due to the good property\,\cite{Hurley_2009_TIT_ComparingSparseMeasures},
we use the Gini index to measure the sparse level of each \emph{abundance}
vector. Given a vector $\mtbfa$ with its elements sorted as $a_{\left(1\right)}\leq a_{\left(2\right)}\leq\cdots\leq a_{\left(K\right)}$,
the Gini index is defined as
\begin{equation}
S\left(\mtbfa\right)=1-2\sum_{k=1}^{K}\frac{a_{\left(k\right)}}{\left\Vert \mtbfa\right\Vert _{1}}\left(\frac{K-k+\frac{1}{2}}{K}\right).\label{eq:definition_Gini_index}
\end{equation}
In this way, large elements have a smaller weight than the small elements
in the sparse measure, avoiding the situation where smaller elements
have negligible (or no) effect on the measure of sparsity\,\cite{Hurley_2009_TIT_ComparingSparseMeasures}.

\subsection{\label{sub:L21_robustModel}Robust HU Model via Joint $\ell_{2,1}$-norm
and $\ell_{p}$-norm }

Considering the robust representation loss\ \eqref{eq:RRLbS_robust_L2p_loss_matrix}
and the pixel-level sparsity constraint\ \eqref{eq:RRLbS_individual_sparseConstraint_matrix}
together, the overall HU objective (RRLbS) is given by
\begin{align}
\min_{\mtbfM\geq\mtbfzero,\mtbfA\geq\mtbfzero} & \mtcalO=\frac{1}{2}\left\Vert \mtbfX-\mtbfM\mtbfA\right\Vert _{2,1}+\lambda\left\Vert \mtbfA^{\mtbfone-\mtbfH}\right\Vert _{1},\label{eq:RRLbS_final_model}
\end{align}
where $\lambda$ is a nonnegative balancing parameter. Due to the
non-convex and non-smooth property of\ \eqref{eq:RRLbS_final_model},
the above objective is very challenging to solve. The efficient  solver
as well as its convergent proofs and computational complexity analyses
will be systematically provided in the next section.

\section{\label{sec:algorithm_for_RRLbS}An Efficient Algorithm for RRLbS
}

Since\ \eqref{eq:RRLbS_final_model} is highly non-convex and non-smooth
in $\mtbfM$ and $\mtbfA$, the final objective\ \eqref{eq:RRLbS_final_model}
is challenging to solve. As a profound theoretical contribution, we
propose an efficient iterative algorithm to solve the joint $\ell_{2,1}$-norm
and $\ell_{p}$-norm based model. We first introduce how to efficiently
solve\ \eqref{eq:RRLbS_final_model} and then give a systematic analysis
to the proposed solver, including its convergence and computation
complexity.

\subsection{Updating Rules for RRLbS}

To ensure the Lipschitz condition\ \cite{JChYe_2012_JNM_Lipschitz,fyzhu_2014_TIP_DgS_NMF}
of the learning-based sparsity constraint, we reformulate our model
as
\begin{equation}
\min_{\mtbfM\geq\mtbfzero,\mtbfA\geq\mtbfzero}\mtcalO=\frac{1}{2}\left\Vert \mtbfX-\mtbfM\mtbfA\right\Vert _{2,1}+\lambda\left\Vert \left(\mtbfA+\xi\right)^{\mtbfone-\mtbfH}\right\Vert _{1},\label{eq:RRLbS_final_model_Lipschitz}
\end{equation}
 where $\xi$ is a small positive value adding to every entry in $\mtbfA$.
As a result, $A_{kn}\!+\!\xi\!>\!0\left(\forall k,n\right)$ guarantees
the Lipschitz condition of the learning-based sparsity constraint.
It is obvious that\ \eqref{eq:RRLbS_final_model_Lipschitz} is reduced
to\ \eqref{eq:RRLbS_final_model} when $\xi\rightarrow0$.

Then,  the Lagrangian multiplier is used to deal with the nonnegative
constraint on $\mtbfM$ and $\mtbfA$, resulting in the following
objective
\begin{align}
\min_{\mtbfM,\mtbfA}\ \mtcalL= & \frac{1}{2}\left\Vert \mtbfX-\mtbfM\mtbfA\right\Vert _{2,1}+\lambda\left\Vert \left(\mtbfA+\xi\right)^{\mtbfone-\mtbfH}\right\Vert _{1}\label{eq:RRLbS_Lagrangian_obj_M_A}\\
 & +\mttrace\left(\Lambda\mtbfM^{\intercal}\right)+\mttrace\left(\Gamma\mtbfA^{T}\right),\nonumber 
\end{align}
where $\Lambda\in\mtbbR_{+}^{L\times K}$ and $\Gamma\in\mtbbR_{+}^{K\times N}$
are the Lagrangian multipliers of the inequality constraints $\mtbfM\geq\mtbfzero$
and $\mtbfA\geq\mtbfzero$ respectively. There are two variable matrices
in\ \eqref{eq:RRLbS_Lagrangian_obj_M_A}. Thus, an alternate algorithm
is proposed. Specifically, the solution with respect to $\left\{ \mtbfM,\mtbfA\right\} $
is given in the following theorem. 
\begin{thm}
 An updated point $\left\{ \mtbfM,\mtbfA\right\} $ could be achieved
via the updating rules as{\small{}
\begin{align}
M_{lk} & \leftarrow M_{lk}\frac{\left(\mtbfU\mtbfX\mtbfA^{T}\right)_{lk}}{\left(\mtbfU\mtbfM\mtbfA\mtbfA^{T}\right)_{lk}}\label{eq:RRLbS_update_M}\\
A_{kn} & \leftarrow A_{kn}\frac{\left(\mtbfM^{T}\mtbfU\mtbfX\right)_{kn}}{\left(\mtbfM^{T}\mtbfU\mtbfM\mtbfA+\lambda\left(1-\mtbfH\right)\circ\left(\mtbfA+\xi\right)^{-\mtbfH}\right)_{kn}},\label{eq:RRLbS_update_A}
\end{align}
}where $\mtbfU\in\mtbbR_{+}^{L\times L}$ is a positive-definite and
diagonal matrix with the $l^{\mtth}$ diagonal entry\footnote{To avoid singular failures, if $\left(\mtbfM\mtbfA-\mtbfX\right)^{l}\!=\!\mtbfzero$,
we obtain $U_{ll}$ by $U_{ll}\!=\!\frac{1}{2}\sqrt{\bigl\Vert\left(\mtbfM\mtbfA-\mtbfX\right)^{l}\bigr\Vert_{2}^{2}+\epsilon}$,
where $\epsilon$ is typically set $10^{-8}$. } as $U_{ll}=\frac{1}{2}\bigl\Vert\left(\mtbfM\mtbfA-\mtbfX\right)^{l}\bigr\Vert_{2}^{-1}$;
$\circ$ is the Hadamard product between matrices; $\mtbfA^{-\mtbfH}\!=\left[A_{kn}^{-H_{kn}}\right]$
is an element-wise exponential operation. \end{thm}
\begin{IEEEproof}
\noindent According to the constrained optimization, a stationary
point of\ \eqref{eq:RRLbS_Lagrangian_obj_M_A} could be achieved
by differentiating\ \eqref{eq:RRLbS_Lagrangian_obj_M_A}, setting
the partial derivatives to zero and considering the Karush-Kuhn-Tucker
(KKT) optimality conditions\ \cite{Nocedal_2006_book_numericalOptimization,Lin_2007_neuralCompation_PGM_NMF}.
This amounts to a two-step strategy. First, setting the partial derivatives
to zero, we have 
\begin{align}
\nabla_{\mtbfM}\mtcalL= & \mtbfU\left(\mtbfM\mtbfA-\mtbfX\right)\mtbfA^{T}+\Lambda=\mtbfzero\label{eq:RRLbS_derivative_M}\\
\nabla_{\mtbfA}\mtcalL= & \mtbfM^{\intercal}\mtbfU\left(\mtbfM\mtbfA-\mtbfX\right)+\Gamma+\nonumber \\
 & \lambda\left(1-\mtbfH\right)\circ\left(\mtbfA+\xi\right)^{-\mtbfH}=\mtbfzero.\label{eq:RRLbS_derivative_A}
\end{align}
Then, considering the KKT conditions $\Lambda_{lk}M_{lk}=0$ and $\Gamma_{kn}A_{kn}=0$,
we have the equalities as 
\begin{align*}
\left(\mtbfU\mtbfM\mtbfA\mtbfA^{T}-\mtbfU\mtbfX\mtbfA^{T}\right)_{lk}M_{lk} & =0\\
\left(\mtbfM^{T}\mtbfU\left(\mtbfM\mtbfA-\mtbfX\right)+\lambda\left(1-\mtbfH\right)\circ\left(\mtbfA+\xi\right)^{-\mtbfH}\right)_{kn}A_{kn} & =0.
\end{align*}
Solving the above equations, we get the final updating rules
\begin{align*}
M_{lk} & \leftarrow M_{lk}\frac{\left(\mtbfU\mtbfX\mtbfA^{T}\right)_{lk}}{\left(\mtbfU\mtbfM\mtbfA\mtbfA^{T}\right)_{lk}}\\
A_{kn} & \leftarrow A_{kn}\frac{\left(\mtbfM^{T}\mtbfU\mtbfX\right)_{kn}}{\left(\mtbfM^{T}\mtbfU\mtbfM\mtbfA+\lambda\left(1-\mtbfH\right)\circ\left(\mtbfA+\xi\right)^{-\mtbfH}\right)_{kn}}.
\end{align*}
In this manner, we give the solver for the objective\ \eqref{eq:RRLbS_final_model_Lipschitz}.
\end{IEEEproof}
\begin{algorithm}[t]
\caption{\textbf{for RRLbS\ \eqref{eq:RRLbS_final_model} \label{alg:DgS_nmf}}}
 \textbf{Input:} hyperspectral image $\mtbfX$; the number of \emph{endmembers}
$K$; parameter $\lambda$; the initial guidance map $\mtbfh^{\left(0\right)}$. 

\begin{algorithmic}[1] 

\STATE initialize\emph{ }the factor matrices $\mtbfM$ and $\mtbfA$.

\STATE calculate $\mtbfH=\mtbfone_{K}\left(\mtbfh^{\left(0\right)}\right)^{\intercal}\in\mtbbR^{K\times N}.$

\REPEAT

\REPEAT

\STATE update $\mtbfA$ via the updating rule~\eqref{eq:RRLbS_update_A}.

\STATE update $\mtbfM$ via the updating rule~\eqref{eq:RRLbS_update_M}.

\UNTIL{stable} 

\STATE update $\mtbfH$ via the updating rule~\eqref{eq:update_H}.

\UNTIL{convergence} 

\end{algorithmic} 

\textbf{Output} $\mtbfM$ and $\mtbfA$ as the final unmixing result.
\end{algorithm}
As mentioned before, the most direct clue to estimate the guidance
map $\mtbfh$ is the crucial dependence upon \emph{abundances.} Once
getting the stable \emph{abundance $\mtbfA$}, $\mtbfh$ could be
efficiently solved via the Gini index\,\eqref{eq:definition_Gini_index}
as 
\begin{equation}
\mtbfH=\mtbfone_{K}\mtbfh^{T},\mtbfh=S\left(\mtbfA\right).\label{eq:update_H}
\end{equation}
Note that, to satisfy the $\ell_{p}$ sparse constraint in\ \eqref{eq:RRLbS_individual_sparseConstraint_vector},
every value in the guidance map needs to be scaled into the range
of $\left[0,0.5\right]$, that is 
\[
h_{n}\leftarrow\frac{h_{n}-\min\left(\mtbfh\right)}{2\left[\max\left(\mtbfh\right)-\min\left(\mtbfh\right)\right]},\quad\forall n\in\left\{ 1,2,\cdots,N\right\} .
\]
The solver for the RRLbS model\,\eqref{eq:RRLbS_final_model} is
given in Algorithm\,\ref{alg:DgS_nmf}.

\subsection{\label{sub:ConvergenceProof_RRLbS}Convergence Analysis }

To ensure the reliability of~\eqref{eq:RRLbS_update_M} and~\eqref{eq:RRLbS_update_A},
we would like to analyze their convergence property.
\begin{lem}
\label{lem:updatingRules_equivalent}The updating rules~\eqref{eq:RRLbS_update_M}
and~\eqref{eq:RRLbS_update_A} are equivalent to the following updating
rules
\begin{align}
\widehat{M}_{lk} & \leftarrow\widehat{M}_{lk}\frac{\left(\mhbfX\mtbfA^{T}\right)_{lk}}{\left(\mhbfM\mtbfA\mtbfA^{T}\right)_{lk}}\label{eq:updataDgS_M}\\
A_{kn} & \leftarrow A_{kn}\frac{\left(\mhbfM^{T}\mhbfX\right)_{kn}}{\left(\mhbfM^{T}\mhbfM\mtbfA+\lambda\left(\mtbfone-\mtbfH\right)\circ\left(\mtbfA+\xi\right)^{-\mtbfH}\right)_{kn}},\label{eq:updateDgS_A}
\end{align}
where $\mhbfM=\mtbfU^{\frac{1}{2}}\mtbfM,\mhbfX=\mtbfU^{\frac{1}{2}}\mtbfX$.
This means that the objective\ \eqref{eq:RRLbS_final_model} could
be equivalently solved by\ \eqref{eq:updataDgS_M} and\ \eqref{eq:updateDgS_A}. \end{lem}
\begin{IEEEproof}
Considering $\mhbfM=\mtbfU^{\frac{1}{2}}\mtbfM,\mhbfX=\mtbfU^{\frac{1}{2}}\mtbfX$
and that $\mtbfU$ is a positive-definite and diagonal matrix, the
updating rules\ \eqref{eq:updataDgS_M} and\ \eqref{eq:updateDgS_A}
become
\begin{equation}
U_{ll}^{\frac{1}{2}}M_{lk}\leftarrow U_{ll}^{\frac{1}{2}}M_{lk}\frac{\left(\mtbfU^{\frac{1}{2}}\mtbfX\mtbfA^{T}\right)_{lk}}{\left(\mtbfU^{\frac{1}{2}}\mtbfM\mtbfA\mtbfA^{T}\right)_{lk}}\label{eq:updataDgS_M-1}
\end{equation}
\begin{align}
A_{kn} & \leftarrow A_{kn}\frac{\left(\mtbfM^{T}\mtbfU\mtbfY\right)_{kn}}{\left(\mtbfM^{T}\mtbfU\mtbfM\mtbfA+\lambda\left(\mtbfone-\mtbfH\right)\circ\left(\mtbfA+\xi\right)^{-\mtbfH}\right)_{kn}}.\label{eq:updateDgS_A-1}
\end{align}
Since $\left(\mtbfU^{\frac{1}{2}}\mtbfY\right)_{lk}=U_{ll}^{\frac{1}{2}}Y_{lk},\forall\mtbfY\in\mtbbR^{L\times K}$,
we have the following derivations
\begin{equation}
\frac{\left(\mtbfU^{\frac{1}{2}}\mtbfX\mtbfA^{T}\right)_{lk}}{\left(\mtbfU^{\frac{1}{2}}\mtbfM\mtbfA\mtbfA^{T}\right)_{lk}}=\frac{U_{ll}^{\frac{1}{2}}\left(\mtbfU^{\frac{1}{2}}\mtbfX\mtbfA^{T}\right)_{lk}}{U_{ll}^{\frac{1}{2}}\left(\mtbfU^{\frac{1}{2}}\mtbfM\mtbfA\mtbfA^{T}\right)_{lk}}=\frac{\left(\mtbfU\mtbfX\mtbfA^{T}\right)_{lk}}{\left(\mtbfU\mtbfM\mtbfA\mtbfA^{T}\right)_{lk}}.\label{eq:equation_derivations}
\end{equation}
Substituting\ \eqref{eq:equation_derivations} into\ \eqref{eq:updataDgS_M-1},
we have 
\begin{equation}
M_{lk}\leftarrow M_{lk}\frac{\left(\mtbfU\mtbfX\mtbfA^{T}\right)_{lk}}{\left(\mtbfU\mtbfM\mtbfA\mtbfA^{T}\right)_{lk}}.\label{eq:updateDgS_A-1-1}
\end{equation}
 Since the updating rules\,\eqref{eq:updateDgS_A-1-1},\,\eqref{eq:updateDgS_A-1}
are exact the same as the updating rules\,\eqref{eq:RRLbS_update_M},\,\eqref{eq:RRLbS_update_A},
we have proved Lemma\ \ref{lem:updatingRules_equivalent}. \end{IEEEproof}
\begin{thm}
\label{thm:nonincreasing_theorem}The objective~\eqref{eq:RRLbS_final_model}
is non-increasing by using the updating rules~\eqref{eq:RRLbS_update_M}
and~\eqref{eq:RRLbS_update_A}.\end{thm}
\begin{IEEEproof}
Based on the partial derivatives\ \eqref{eq:RRLbS_derivative_M},\,\eqref{eq:RRLbS_derivative_A},
it is obvious that the updating rules~\eqref{eq:RRLbS_update_M},\,\eqref{eq:RRLbS_update_A}
compute the optimal solution of the following problem, 
\begin{align}
\min_{\mtbfM\geq\mtbfzero,\mtbfA\geq\mtbfzero}\, & \frac{1}{2}\mttrace\left\{ \left(\mtbfX-\mtbfM\mtbfA\right)^{T}\mtbfU\left(\mtbfX-\mtbfM\mtbfA\right)\right\} \nonumber \\
 & +\lambda\left\Vert \left(\mtbfA+\xi\right)^{1-\mtbfH}\right\Vert _{1}\label{eq:proof_our_obj_1}
\end{align}
which is equivalent to the objective as:
\begin{equation}
\min_{\mhbfM\geq\mtbfzero,\mtbfA\geq\mtbfzero}\frac{1}{2}\left\Vert \mhbfX-\mhbfM\mtbfA\right\Vert _{F}^{2}+\lambda\left\Vert \left(\mtbfA+\xi\right)^{1-\mtbfH}\right\Vert _{1},\label{eq:proof_our_obj_2}
\end{equation}
where $\mhbfM=\mtbfU^{\frac{1}{2}}\mtbfM,\mhbfX=\mtbfU^{\frac{1}{2}}\mtbfX$.
It has been proved in\ \cite{fyzhu_2014_TIP_DgS_NMF} that\ \eqref{eq:proof_our_obj_2}
is non-increasing under the updating rules\ \eqref{eq:updataDgS_M},\,\eqref{eq:updateDgS_A}.
Because\ \eqref{eq:proof_our_obj_1} is the same problem as\ \eqref{eq:proof_our_obj_2},
and the updating rules~\eqref{eq:RRLbS_update_M},\,\eqref{eq:RRLbS_update_A}
are equivalent to\ \eqref{eq:updataDgS_M},\,\eqref{eq:updateDgS_A},
as analyzed in Lemma\ \ref{lem:updatingRules_equivalent}, we infer
that the objective\ \eqref{eq:proof_our_obj_1} is non-increasing
under the updating rules~\eqref{eq:RRLbS_update_M},\,\eqref{eq:RRLbS_update_A}
as well. For each iteration, we denote the updated $\left\{ \mtbfM,\mtbfA\right\} $
as $\left\{ \mebfM,\mebfA\right\} $. Thus, we have the following
inequalities
\begin{align*}
 & \frac{1}{2}\mttrace\left\{ \mebfE^{T}\mtbfU\mebfE\right\} +\lambda\left\Vert \left(\mebfA+\xi\right)^{1-\mtbfH}\right\Vert _{1}\\
\leq & \frac{1}{2}\mttrace\left\{ \mtbfE^{T}\mtbfU\mtbfE\right\} +\lambda\left\Vert \left(\mtbfA+\xi\right)^{1-\mtbfH}\right\Vert _{1},
\end{align*}
where $\mebfE=\mtbfX-\mebfM\mebfA$ and $\mtbfE=\mtbfX-\mtbfM\mtbfA$.
This amounts to 
\begin{align}
 & \sum_{l=1}^{L}\frac{\left\Vert \mebfe^{l}\right\Vert _{2}^{2}}{2\left\Vert \mtbfe^{l}\right\Vert _{2}}+\lambda\left\Vert \left(\mebfA+\xi\right)^{1-\mtbfH}\right\Vert _{1}\nonumber \\
\leq & \sum_{l=1}^{L}\frac{\left\Vert \mtbfe^{l}\right\Vert _{2}^{2}}{2\left\Vert \mtbfe^{l}\right\Vert _{2}}+\lambda\left\Vert \left(\mtbfA+\xi\right)^{1-\mtbfH}\right\Vert _{1}.\label{eq:auxiliaryObj_inequality}
\end{align}
Given the function ${\displaystyle f\left(x\right)=x-\frac{x^{2}}{2\alpha}}\left(\forall\alpha\in\mtbbR_{+}\right)$,
$f\left(x\right)\leq f\left(\alpha\right)$ holds for any $x\in\mtbbR$\ \cite{huaWang_2014_ICML_RobustMetricLearning,FpNie_2013_IJCAI_EarlyActiveLearning}.
Thus we have
\begin{equation}
\sum_{l=1}^{L}\left\Vert \mebfe^{l}\right\Vert _{2}-\sum_{l=1}^{L}\frac{\left\Vert \mebfe^{l}\right\Vert _{2}^{2}}{2\left\Vert \mtbfe^{l}\right\Vert _{2}}\leq\sum_{l=1}^{L}\left\Vert \mtbfe^{l}\right\Vert _{2}-\sum_{l=1}^{L}\frac{\left\Vert \mtbfe^{l}\right\Vert _{2}^{2}}{2\left\Vert \mtbfe^{l}\right\Vert _{2}}.\label{eq:auxiliary_inequality_simpleFunc}
\end{equation}
Combining the inequalities of\ \eqref{eq:auxiliaryObj_inequality}
and\eqref{eq:auxiliary_inequality_simpleFunc} together, we have the
inequality as 
\begin{align*}
 & \sum_{l}\left\Vert \mebfe^{l}\right\Vert _{2}+\lambda\left\Vert \left(\mebfA+\xi\right)^{1-\mtbfH}\right\Vert _{1}\\
\leq & \sum_{l}\left\Vert \mtbfe^{l}\right\Vert _{2}+\lambda\left\Vert \left(\mtbfA+\xi\right)^{1-\mtbfH}\right\Vert _{1},
\end{align*}
which is equivalent to the inequality 
\begin{align*}
 & \frac{1}{2}\left\Vert \mtbfX-\mebfM\mebfA\right\Vert _{2,1}+\lambda\left\Vert \left(\mebfA+\xi\right)^{\mtbfone-\mtbfH}\right\Vert _{1}\\
\leq & \frac{1}{2}\left\Vert \mtbfX-\mtbfM\mtbfA\right\Vert _{2,1}+\lambda\left\Vert \left(\mtbfA+\xi\right)^{\mtbfone-\mtbfH}\right\Vert _{1}.
\end{align*}
In this way, we have proven Theorem\ \ref{thm:nonincreasing_theorem}.
\end{IEEEproof}
Apart from the theoretical proof above, the empirical convergent study
for Algorithm\ \ref{alg:DgS_nmf} is summarized in Section\ \ref{sub:Convergence-Study}.
\begin{table*}[t]
\centering{}\caption{Computational operation counts for NMF and RRLbS at each iteration.\label{tab:ComplexSummarize}}
\begin{tabular}{|r|cccc|c|}
\hline 
\multirow{2}{*}{Methods} &
\multicolumn{4}{c|}{Arithmetic Operations in float-point format at each iteration } &
\multirow{2}{*}{Overall}\tabularnewline
\cline{2-5} 
 & Addition &
Multiplication &
Division &
Exponent & \tabularnewline
\hline 
\multirow{2}{*}{NMF} &
$2LNK-2K\left(N+L\right)$ &
$2LNK+K\left(L+N\right)$ &
\multirow{2}{*}{$K\left(L+N\right)$} &
\multirow{2}{*}{--} &
\multirow{2}{*}{$O\left(KLN\right)$}\tabularnewline
 & $+2K^{2}\left(L+N\right)-2K^{2}$ &
 $+2K^{2}\left(L+N\right)$ &  &  & \tabularnewline
\hline 
\multirow{2}{*}{RRLbS} &
$3KLN-K\left(L+N\right)$ &
$3KLN+L\left(3K+N\right)$ &
\multirow{2}{*}{$K\left(L+N\right)$} &
\multirow{2}{*}{$KN+L$} &
\multirow{2}{*}{$O\left(KLN\right)$}\tabularnewline
 & $+2K^{2}\left(L+N\right)-2K^{2}$ &
$+K^{2}\left(2L+3N\right)$ &  &  & \tabularnewline
\hline 
\end{tabular}
\end{table*}
\begin{table}[t]
\centering{}\caption{Parameters used in Computational Complexity Analysis. \label{tab:complexParameters}}
\begin{tabular}{|c|l|}
\hline 
Parameters &
Description\tabularnewline
\hline 
$K$ &
number of \emph{endmembers}\tabularnewline
\hline 
 $L$ &
number of channels\tabularnewline
\hline 
$N$ &
number of pixels in hyperspectral image\tabularnewline
\hline 
$t$ &
number of iteration steps\tabularnewline
\hline 
\end{tabular}
\end{table}

\subsection{\label{sub:Computational-Complexity-Analysi}Computational Complexity
Analysis}

Theoretically, the computational complexity is important for algorithms.
In this section, we analyze the additional computational cost of our
method compared with the standard NMF. To give a precise comparison,
the arithmetic operations of addition, multiplication, division and
exponent, are counted for each iteration. 

Based on the updating rules~\eqref{eq:RRLbS_update_M} and~\eqref{eq:RRLbS_update_A},
it is easy to summarize the counts of operations in Tabel\ \ref{tab:ComplexSummarize},
where the notations are listed in Table\ \ref{tab:complexParameters}.
For RRLbS, it is important to note that $\mtbfU$ is a positive-defined
diagonal matrix. This property facilitates the savage of computational
costs. For example, it only costs $L$ exponent operations to compute
the exponent of $\mtbfU\!\in\!\mtbbR_{+}^{L\times L}$, that is $\mtbfU^{\alpha}\!=\!\left\{ U_{ll}^{\alpha}\right\} _{l=1}^{L},\forall\alpha\!\in\!\mtbbR$.
While for a normal matrix $\mtbfV\in\mtbbR_{+}^{L\times L}$ of the
same size, it costs $O\left(L^{3}\right)$ to get the inverse matrix
$\mtbfV^{\alpha}\left(\alpha\!=\!-\!1\right)$, which could also be
treated as an exponent operation. The cost of matrix multiplication
$\mtbfU\mtbfM$ is greatly saved as well; it costs $LK$ multiplication
for our case. While for a normal $\mtbfV\in\mtbbR_{+}^{L\times L}$,
it takes $L^{2}K$ multiplication and $L^{2}K-LK$ addition to get
$\mtbfV\mtbfM$. 

Apart from the updating costs, RRLbS costs $O\left(4LN\right)$ to
get the initial guidance map and $O\left(NK+NK\log K\right)$ to get
an updated one. If the updating process stops after $t$ steps and
the learning-based guidance map updates after each $q$ (typically
$10$) iterations, the total cost of RRLbS is 
\[
O\left(tKLN+4LN+\frac{t}{q}\left(NK+NK\log K\right)\right).
\]
While the total cost of NMF is $O\left(tKLN\right)$. Generally $N\gg\max\left\{ L,K\right\} ,$
the computational complexities of RRLbS and NMF are of the same magnitude.

\section{\label{sec:Evaluation}Evaluation}

In this section, extensive experiments are conducted to evaluate
the effectiveness of our method in the HU task. 
\begin{figure}[t]
\begin{centering}
\subfloat[{\small{}Urban }\label{fig:realHyperImages_urban}]{\begin{centering}
\includegraphics[width=0.35\columnwidth]{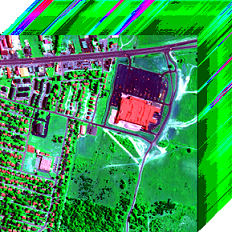}
\par\end{centering}

\centering{}}\subfloat[{\small{}Jasper Ridge }\label{fig:realHyperImages_japser}]{\begin{centering}
\includegraphics[width=0.275\columnwidth]{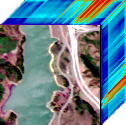}
\par\end{centering}

\centering{}}\subfloat[{\small{}Cuprite }\label{fig:realHyperImages_Cuprite}]{\begin{centering}
\includegraphics[width=0.25\columnwidth]{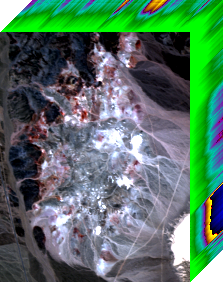}
\par\end{centering}

\centering{}}
\par\end{centering}

\centering{}\caption{Three benchmark hyperspectral images, that is Urban, Jasper Ridge
and Cuprite, used in the experiments.  \label{fig:3_realHyperImages}}
\end{figure}

\subsection{Datasets}

Three real hyperspectral datasets are used in the experiments. Their
information is listed below. The ground truths are obtained by the
method introduced in\ \cite{fyzhu_2014_GroundTruth4HU,fyzhu_2014_IJPRS_SSNMF,SJia_07_TGRS_SSCBSS,fyzhu_2017_HyperDataSets}. 

\textbf{Urban} is one of the most widely used datasets for the HU
studying~\cite{LiuXueSong_2011_TGRS_ConstrainedNMF,Qian_11_TGRS_NMF+l1/2,Jia_09_TGRS_ConstainedNMF}.
There are $307\times307$ pixels. Each pixel is recorded at $210$
channels ranging from $400\,nm$ to $2500\,nm$. Due to the dense
water vapor and the atmospheric effects etc., the channels of $1$--$4$,
$76$, $87$, $101$--$111$, $136$--$153$ and $198$--$210$ are
either blank or badly noised, which, however, are kept for the unmixing
of the hyperspectral image. There are four \emph{endmembers} in this
image: ``\#1 Asphalt'', ``\#2 Grass'', ``\#3 Tree'' and ``\#4 Roof''
as shown in Fig.~\ref{fig:realHyperImages_urban}.

\textbf{Jasper Ridge, }as shown in Fig.~\ref{fig:realHyperImages_japser},\textbf{
}is a popular dataset used in~\cite{rodarmel2002principal,vargas2015colored,tong2016nonnegative,shu2015multilayer,tong2017region,vasuki2017clustering,ganesanmaximin,fu2015low,yingWang_2015_TIP_RobustUnmixing,aggarwal2016hyperspectral}.
It consists of $512\times614$ pixels; each pixel is recorded at $224$
channels ranging from $380\,nm$ to $2500\,nm$, resulting in a very
high spectral resolution as $9.46\,nm$. Since this image is highly
complex, we consider a sub-image of $100\times100$ pixels. This
sub-image starts from the $\left(105,269\right)$-th pixel. Due to
dense water vapor and atmospheric effects etc., the channels $1$--$3$,
$108$--$112$, $154$--$166$ and $220$--$224$ are blank or badly
noised, which, however, are kept for the unmixing process. There
are four \emph{endmembers}, that is  ``\#1 Tree'', ``\#2 Soil'', ``\#3
Water'' and ``\#4 Road'' respectively. 
\begin{figure*}[t]
\begin{centering}
\includegraphics[width=2.05\columnwidth]{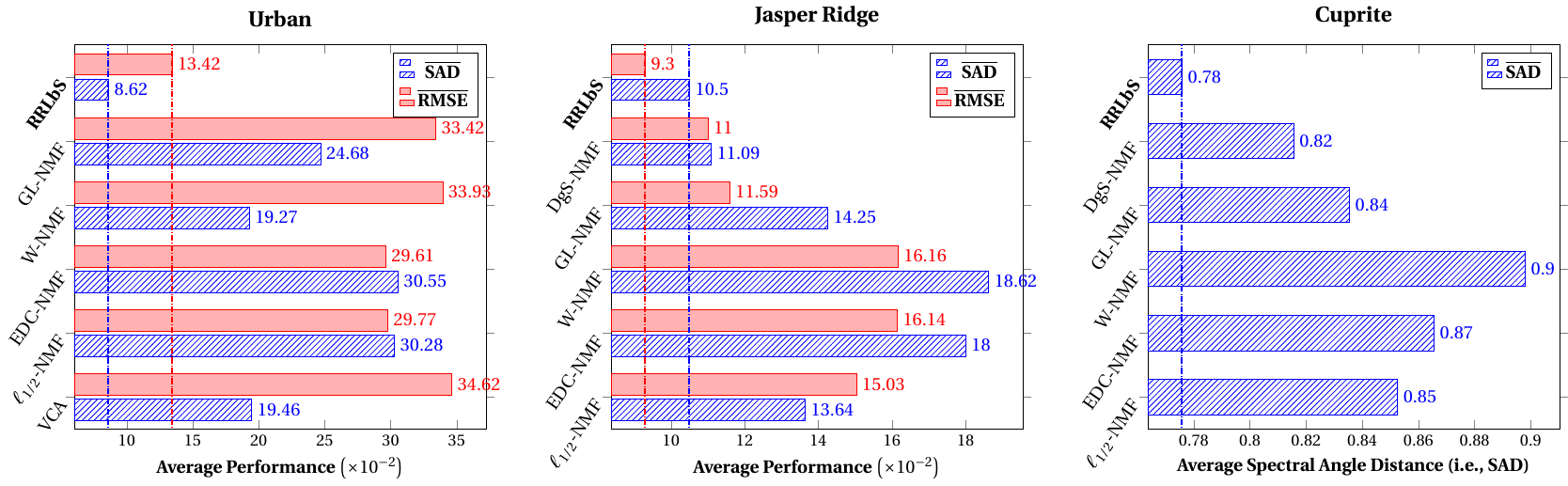}
\par\end{centering}

\centering{}\caption{The average performances (i.e. $\overline{\text{SAD}}$ and $\overline{\text{RMSE}}$)
of six methods on (a) Urban (b) Jasper Ridge and (c) Cuprite. Compared
with the state-of-the-art method, our method achieves highly promising
HU results. \label{fig:AveragedPerformanceOn3DataSets}}
\end{figure*}

\textbf{Cuprite} (cf. Fig.~\ref{fig:realHyperImages_Cuprite}) is
the most benchmark hyperspectral image for the HU research\ \cite{agathos2014robust,wei2017unsupervised,bernabe2017parallel,wang2017spatial,wang2016hypergraph,martel2017gpu,fyzhu_2014_TIP_DgS_NMF,XLu_2013_TGRS_ManifoldSparseNMF,LiuXueSong_2011_TGRS_ConstrainedNMF,Qian_11_TGRS_NMF+l1/2,nWang_13_SelectedTopics_EDC-NMF,Jose_05_TGRS_Vca}.
It is captured by the AVIRIS sensor that covers a Cuprite area in
Las Vegas, NV, U.S. There are 224 spectral bands that range the spectra
from $370\thinspace nm$ to $2,480\thinspace nm$. The bands 1–2,
104–113, 148–167 and 221–224 are noisy bands or water absorption bands,
which are kept for the unmixing process.

In this paper, a subimage of $250\times190$ pixels is considered,
which is widely used in the state-of-the-art HU papers\ \cite{Jose_05_TGRS_Vca,Qian_11_TGRS_NMF+l1/2,XLu_2013_TGRS_ManifoldSparseNMF,fyzhu_2014_TIP_DgS_NMF}.
The researchers have different opinions on the number of endmembers.
In\ \cite{Jose_05_TGRS_Vca}, there are 14 endmembers; while there
are 10 endmembers in\ \cite{Qian_11_TGRS_NMF+l1/2}; then Dr. Lu
hold that there are 12 endmembers in the Cuprite. In this paper, we
agree with Dr. Lu's setting. Please refer to\ \cite{fyzhu_2017_HyperDataSets}
for the illustration of the 12 endmembers. Due to the different setting
of endmembers, the results of the state-of-the-art methods are different
in the papers\ \cite{XLu_2013_TGRS_ManifoldSparseNMF,LiuXueSong_2011_TGRS_ConstrainedNMF,Qian_11_TGRS_NMF+l1/2,nWang_13_SelectedTopics_EDC-NMF,Jose_05_TGRS_Vca}.

\subsection{Compared Algorithms and Parameter Settings}

To verify the superior performance, the proposed method is compared
with six state-of-the-art methods. The details of all these methods
(including our method) are listed as follows: 
\begin{enumerate}
\item \textbf{Our method}: Effective Spectral Unmixing via Robust Representation
and Learning-based Sparsity (RRLbS) is a new method proposed in this
paper. 
\item Vertex Component Analysis~\cite{Jose_05_TGRS_Vca} (VCA) is the benchmark
geometric method. The code is available on the webpage \url{http://www.lx.it.pt/bioucas/code.htm}.
\item $\ell_{1/2}$ sparsity-constrained NMF\,\cite{Qian_11_TGRS_NMF+l1/2}
($\ell{}_{\text{1/2}}$-NMF) is a state-of-the-art method that could
get sparser results than $\ell_{1}$-NMF. Since the code is unavailable,
we implement it.
\item Local Neighborhood Weights regularized NMF~\cite{JmLiu_12_SlectedTopics_W-NMF}
(W-NMF) is a manifold graph based NMF method.  It integrates the
spectral information and spatial information when constructing the
weighted graph. Since this work is an extension of G-NMF\ \cite{Cai_11_PAMI_GNMF},
we implement the code by referring to the code on\ \url{http://www.cad.zju.edu.cn/home/dengcai/Data/GNMF.html}. 
\item \emph{Endmember} Dissimilarity Constrained NMF~\cite{nWang_13_SelectedTopics_EDC-NMF}
(EDC-NMF)  urges the \emph{endmember} to be smooth itself and different
from each other. We implement the code since the orginal code is not
available on the web
\item Graph-regularized $\ell_{1/2}$-NMF~\cite{XLu_2013_TGRS_ManifoldSparseNMF}
(GL-NMF) is a new method proposed in 2013. It considers both the sparse
characteristic and the intrinsic manifold structure in hyperspectral
images. We implement the code by ourself.
\item Data-guided sparsity constrainted NMF\ \cite{fyzhu_2014_TIP_DgS_NMF}
(DgS-NMF) is a state-of-the-art method published in 2014. The main
idea is to apply the adaptively sparse constraints, which are according
to the mixed level of each pixel.
\end{enumerate}
There is no parameter in VCA. For the other six methods, there is
one main parameter. To find a good parameter setting, typical procedures
consist of two phases: a bracketing phase that finds an interval $\left[\lambda_{\min},\,\lambda_{\max}\right]$
containing acceptable parameters, and a selection phase that zooms
in to locate the optimal parameter.

\subsection{Evaluation Metrics for Quantitative Performances}

We use two benchmark metrics to measure the quantitative HU results,
i.e. (a) Spectral Angle Distance (SAD)~\cite{LiuXueSong_2011_TGRS_ConstrainedNMF,Jose_05_TGRS_Vca}
and (b) Root Mean Square Error (RMSE)~\cite{LiuXueSong_2011_TGRS_ConstrainedNMF,Qian_11_TGRS_NMF+l1/2,Kelly_2011_TGRS_SpatiAdaptiveUnmixing}.
 Both metrics assess the estimated errors. Thus, the smaller SAD
and RMSE correspond to the better results. Specifically, SAD evaluates
the estimated \emph{endmember}, and RMSE assesses the estimated \emph{abundance}
map. They are defined as 
\begin{equation}
\mbox{SAD}\left(\mtbfm_{k},\mhbfm_{k}\right)=\arccos\left(\frac{\mtbfm_{k}^{T}\mhbfm_{k}}{\|\mtbfm_{k}\|\cdot\|\mhbfm_{k}\|}\right)\label{eq:sadMetri_experiment}
\end{equation}
and
\begin{equation}
\mbox{RMSE}\left(\mtbfa^{k},\mhbfa^{k}\right)=\left(\frac{1}{N}\|\mtbfa^{k}-\mhbfa^{k}\|_{2}^{2}\right)^{1/2}\label{eq:rmseMetri_experiment}
\end{equation}
$\forall k=\left\{ 1,\cdots,K\right\} $, where $\mhbfm_{k}$ is the
$k^{\mtth}$ estimated \emph{endmember}, $\mhbfa^{k}$ is the $k^{\mtth}$
estimated \emph{abundance} map (i.e. the $k^{\mtth}$ row vector in
$\mhbfA$), $\left\{ \mtbfm_{k},\mtbfa^{k}\right\} $ are the corresponding
ground truth; $N$ is the number of pixels in image.   
\begin{table*}[t]
\noindent \begin{centering}
\caption{Unmixing performances of seven state-of-the-art methods on Urban.
The \textcolor{red}{red value} is the best, and the \textcolor{blue}{blue
value} is the $2^{\text{nd}}$ best. \label{tab:urban_SAD=000026RMSE}}
\vspace{-0.2cm}
 \resizebox{1.82\columnwidth}{!}{%
\begin{tabular}{|l|ccccccc|}
\hline 
End- &
\multicolumn{7}{c|}{Spectral Angle Distance \textbf{SAD $\left(\times10^{-2}\right)$}}\tabularnewline
\cline{2-8} 
members &
VCA &
$\ell_{1/2}$-NMF &
EDC-NMF &
W-NMF &
\multirow{1}{*}{GL-NMF} &
DgS-NMF &
RRLbS\tabularnewline
\hline 
\#1 Asphalt &
\textcolor{red}{12.22$\pm$3.63} &
24.92$\pm$0.31  &
17.39$\pm$0.41  &
\textcolor{blue}{16.10$\pm$0.47}  &
24.16$\pm$0.71  &
29.58$\pm$0.83 &
17.35$\pm$0.18 \tabularnewline
\#2 Grass &
42.09$\pm$6.93 &
\textcolor{blue}{29.26$\pm$0.11} &
43.48$\pm$0.06  &
43.67$\pm$0.07 &
30.45$\pm$0.08  &
32.64$\pm$0.10 &
\textcolor{red}{8.01$\pm$0.07}\tabularnewline
\#3 Tree &
12.58$\pm$1.04 &
7.45$\pm$0.06 &
9.78$\pm$0.11 &
11.94$\pm$0.23  &
9.77$\pm$0.53 &
\textcolor{blue}{6.94$\pm$0.04} &
\textcolor{red}{3.74$\pm$0.06}\tabularnewline
\#4 Roof &
43.81$\pm$17.65 &
16.19$\pm$0.23 &
50.45$\pm$0.22 &
50.49$\pm$0.28 &
\textcolor{blue}{12.71$\pm$0.20 } &
29.56$\pm$0.61 &
\textcolor{red}{5.38$\pm$0.06}\tabularnewline
Avg. &
27.68 &
19.46  &
30.28 &
30.55 &
\textcolor{blue}{19.27} &
24.68 &
\textcolor{red}{8.62}\tabularnewline
\hline 
\hline 
 &
\multicolumn{7}{c|}{Root Mean Square Error \textbf{RMSE $\left(\times10^{-2}\right)$}}\tabularnewline
\hline 
\#1 Asphalt &
32.60$\pm$1.93  &
29.76$\pm$ 0.12  &
24.12$\pm$0.12  &
\textcolor{blue}{23.84$\pm$0.12 } &
29.19$\pm$0.17  &
30.18$\pm$0.12 &
\textcolor{red}{10.34$\pm$0.05 }\tabularnewline
\#2 Grass &
39.78$\pm$3.39  &
42.12$\pm$0.05 &
34.78$\pm$0.07 &
\textcolor{blue}{34.34$\pm$0.07} &
41.58$\pm$0.07 &
41.73$\pm$0.10 &
\textcolor{red}{18.23$\pm$0.09}\tabularnewline
\#3 Tree &
\textcolor{blue}{33.05$\pm$5.09} &
40.73$\pm$0.03  &
38.82$\pm$0.01 &
38.63$\pm$0.02 &
40.27$\pm$0.02  &
40.72$\pm$0.05 &
\textcolor{red}{13.90$\pm$0.10 }\tabularnewline
\#4 Roof &
33.01$\pm$5.61 &
25.86$\pm$0.39 &
21.36$\pm$0.15 &
21.63$\pm$0.17 &
24.69$\pm$0.59  &
\textcolor{blue}{21.07$\pm$0.30} &
\textcolor{red}{11.20$\pm$0.10}\tabularnewline
Avg. &
34.61 &
34.62 &
29.77 &
\textcolor{blue}{29.61} &
33.93 &
33.42 &
\textcolor{red}{13.42}\tabularnewline
\hline 
\end{tabular}}
\par\end{centering}

\noindent \begin{centering}
\vspace{0.15cm}
\caption{Unmixing performances of seven methods on Jasper Ridge. The \textcolor{red}{red
value} is the best, and the \textcolor{blue}{blue value} is the $2^{\text{nd}}$
best.\label{tab:jasper_SAD=000026RMSE}}
\vspace{-0.2cm}
 \resizebox{1.82\columnwidth}{!}{%
\begin{tabular}{|l|ccccccc|}
\hline 
End- &
\multicolumn{7}{c|}{Spectral Angle Distance \textbf{SAD $\left(\times10^{-2}\right)$}}\tabularnewline
\cline{2-8} 
members &
VCA &
$\ell_{1/2}$-NMF &
EDC-NMF &
W-NMF &
GL-NMF &
DgS-NMF &
\multirow{1}{*}{RRLbS}\tabularnewline
\hline 
\#1 Tree &
24.71$\pm$5.83  &
11.66$\pm$0.14  &
18.21$\pm$0.06  &
18.71$\pm$0.09  &
\textcolor{blue}{9.71$\pm$0.32} &
9.76$\pm$0.08 &
\textcolor{red}{8.71$\pm$0.13 }\tabularnewline
\#2 Soil &
24.72$\pm$0.33 &
18.22$\pm$0.35 &
21.83$\pm$ 0.44 &
22.12$\pm$0.42  &
\textcolor{blue}{10.23$\pm$0.09 } &
12.34$\pm$0.08 &
\textcolor{red}{9.79$\pm$0.14 }\tabularnewline
\#3 Water &
23.68$\pm$14.66  &
\textcolor{red}{3.58$\pm$1.02} &
8.51$\pm$0.53  &
9.99$\pm$0.44 &
20.27$\pm$0.63  &
6.25$\pm$0.27 &
\textcolor{blue}{5.37$\pm$0.38}\tabularnewline
\#4 Road &
52.26$\pm$6.93 &
21.12$\pm$0.11 &
23.44$\pm$0.22 &
23.66$\pm$0.29  &
\textcolor{blue}{16.80$\pm$0.29} &
\textcolor{red}{15.98$\pm$0.22} &
18.13$\pm$0.17 \tabularnewline
Avg. &
31.34 &
13.64 &
18.00  &
18.62 &
14.25  &
\textcolor{blue}{11.09} &
\textcolor{red}{10.50} \tabularnewline
\hline 
\hline 
 &
\multicolumn{7}{c|}{Root Mean Square Error \textbf{RMSE $\left(\times10^{-2}\right)$}}\tabularnewline
\hline 
\#1 Tree &
29.05$\pm$5.90 &
8.23$\pm$0.28 &
13.32$\pm$0.13  &
13.73$\pm$0.07  &
\textcolor{red}{6.57$\pm$0.15}  &
7.76$\pm$0.23 &
\textcolor{blue}{7.04$\pm$0.36 }\tabularnewline
\#2 Soil &
28.26$\pm$9.27 &
6.49$\pm$0.02 &
9.39$\pm$0.08  &
10.22$\pm$0.06  &
\textcolor{blue}{6.12$\pm$0.03}  &
6.47$\pm$0.03 &
\textcolor{red}{5.66$\pm$0.04 }\tabularnewline
\#3 Water &
25.37$\pm$7.85 &
22.88$\pm$0.28 &
21.32$\pm$0.23  &
20.89$\pm$0.28  &
17.73$\pm$0.61 &
\textcolor{blue}{15.73$\pm$0.31} &
\textcolor{red}{13.53$\pm$0.32 }\tabularnewline
\#4 Road &
24.12$\pm$16.81 &
22.52$\pm$0.47 &
20.54$\pm$0.34 &
19.82$\pm$0.44 &
15.94$\pm$0.65  &
\textcolor{blue}{14.06$\pm$0.36} &
\textcolor{red}{10.98$\pm$0.33 }\tabularnewline
Avg. &
26.70 &
15.03 &
16.14 &
16.16  &
11.59 &
\textcolor{blue}{11.00} &
\textcolor{red}{9.30}\tabularnewline
\hline 
\end{tabular}}
\par\end{centering}

\noindent \begin{centering}
\vspace{0.15cm}

\par\end{centering}

\noindent \centering{}\caption{Unmixing performance of six state-of-the-art methods on Cuprite. The
\textcolor{red}{red value} is the best, and the \textcolor{blue}{blue
value} is the $2^{\text{nd}}$ best.\label{tab:Cuprite_SAD}}
\vspace{-0.2cm}
\resizebox{1.99\columnwidth}{!}{%
\begin{tabular}{|l||c|c|c|c|c|c|}
\hline 
\multirow{2}{*}{Endmembers} &
\multicolumn{6}{c|}{Spectral Angle Distance (\textbf{SAD})}\tabularnewline
\cline{2-7} 
 & $\ell_{1/2}$-NMF &
EDC-NMF &
W-NMF &
GL-NMF &
DgS-NMF &
RRLbS\tabularnewline
\hline 
\#1 Alunite &
0.3378$\pm$0.2846 &
0.3274$\pm$0.0085 &
1.3462$\pm$0.0162 &
0.3176$\pm$0.0362 &
\textcolor{red}{0.3000$\pm$0.0876} &
\textcolor{blue}{0.3120$\pm$0.0450}\tabularnewline
\#2 Andradite &
1.3552$\pm$0.0355 &
1.5106$\pm$0.1238 &
1.3972$\pm$0.0474 &
\textcolor{red}{0.8033$\pm$0.0305} &
\textcolor{blue}{0.8797$\pm$0.0516} &
1.4097$\pm$0.0413\tabularnewline
\#3 Buddingtonite &
\textcolor{blue}{1.3189$\pm$0.0350} &
1.3566$\pm$0.0617 &
1.4359$\pm$0.0394 &
1.3863$\pm$0.0970 &
1.3497$\pm$0.0635 &
\textcolor{red}{1.2886$\pm$0.0741}\tabularnewline
\#4 Dumortierite &
0.6131$\pm$0.1236 &
0.6813$\pm$0.4101 &
1.1769$\pm$0.0759 &
1.0918$\pm$0.3249 &
\textcolor{blue}{0.4060$\pm$0.1326} &
\textcolor{red}{0.3342$\pm$0.0851}\tabularnewline
\#5 Kaolinite$_{1}$ &
\textcolor{blue}{0.4148$\pm$0.3163} &
\textcolor{red}{0.3300$\pm$0.3841} &
1.3173$\pm$0.3309 &
0.7197$\pm$0.3104 &
1.2451$\pm$0.4480 &
0.4651$\pm$0.4010\tabularnewline
\#6 Kaolinite$_{2}$ &
0.3216$\pm$0.1869 &
0.4076$\pm$0.3442 &
\textcolor{blue}{0.3097$\pm$0.0567} &
0.3631$\pm$0.4211 &
0.3206$\pm$0.0081 &
\textcolor{red}{0.2861$\pm$0.3214}\tabularnewline
\#7 Muscovite &
1.4228$\pm$0.0226 &
\textcolor{blue}{1.3684$\pm$0.0328} &
1.4419$\pm$0.0263 &
1.4412$\pm$0.0214 &
1.4152$\pm$0.1141 &
\textcolor{red}{1.0391$\pm$0.0665}\tabularnewline
\multicolumn{1}{|l||}{\#8 Montmorillonite} &
1.1161$\pm$0.4560 &
1.1617$\pm$0.3705 &
\textcolor{red}{0.3336$\pm$0.4429} &
1.4234$\pm$0.3953 &
1.1804$\pm$0.4017 &
\textcolor{blue}{0.3479$\pm$0.4526}\tabularnewline
\#9 Nontronite &
\textcolor{blue}{0.2892$\pm$0.0044} &
0.3103$\pm$0.0312 &
0.3024$\pm$0.0136 &
0.2896$\pm$0.0100 &
0.3030$\pm$0.0123 &
\textcolor{red}{0.2825$\pm$0.0182}\tabularnewline
\#10 Pyrope &
1.2089$\pm$0.3985 &
1.3712$\pm$0.1130 &
0.9347$\pm$0.3493 &
\textcolor{red}{0.3543$\pm$0.3791} &
\textcolor{blue}{0.5711$\pm$0.0643} &
1.2644$\pm$0.3729\tabularnewline
\#11 Sphene &
0.3998$\pm$0.4201 &
1.1496$\pm$0.3551 &
0.4702$\pm$0.2028 &
\textcolor{blue}{0.3805$\pm$0.1989} &
\textcolor{red}{0.3614$\pm$0.1437} &
0.8332$\pm$0.3182\tabularnewline
\#12 Chalcedony &
1.4315$\pm$0.1222 &
\textcolor{blue}{0.4117$\pm$0.0607} &
\textcolor{red}{0.3105$\pm$0.0389} &
1.4540$\pm$0.5391 &
1.4538$\pm$0.4531 &
1.4468$\pm$0.4611\tabularnewline
Avg. &
0.8525  &
0.8655 &
0.8980  &
0.8354 &
\textcolor{blue}{0.8155} &
\textcolor{red}{0.7758}\tabularnewline
\hline 
\end{tabular}}\vspace{-0.2cm}
\end{table*}

\subsection{Quantitative Performance Comparisons}

To verify the performance of our method, we conduct the experiments
on three benchmark datasets, i.e., Urban, Jasper Ridge and Cuprite,
as shown in Fig.\,\ref{fig:3_realHyperImages}. Each experiment is
repeated eight times to ensure a reliable comparison. 

The quantitative results are summarized in Tables~\ref{tab:urban_SAD=000026RMSE},~\ref{tab:jasper_SAD=000026RMSE},~\ref{tab:Cuprite_SAD}
and plotted in Fig.~\ref{fig:AveragedPerformanceOn3DataSets}. Specifically,
Table~\ref{tab:urban_SAD=000026RMSE} summarizes the unmixing results
on Urban, where the top sub-table illustrates SADs and the bottom
sub-table shows RMSEs. In each sub-table, each row shows the results
of one \emph{endmembers}, i.e.  ``\#1 Tree'', ``\#2 Soil'', ``\#3
Water'' and ``\#4 Road'' respectively; the last row shows the average
results over the four \emph{endmembers}. As Table~\ref{tab:urban_SAD=000026RMSE}
shows, our method generally achieves the best results. This case is
better illustrated in Fig.~\ref{fig:AveragedPerformanceOn3DataSets}a,
where the average results are illustrated. As we shall see, RRLbS
performs the best---compared with the second best results, our method
reduces $55.27\%$ for $\overline{\text{SAD}}$ and $54.68\%$ for
$\overline{\text{RMSE}}$. Such extraordinary improvements rely on
two reasons. First, due to the atmospheric effects, there are $48$
channels either blank or badly noised. Accordingly, our method is
robust to these side channels and relieves their bad effects on the
unmixing process. Besides, with the help of the guidance map, RRLbS
exploits an individually  sparse constraint according to the mixed
level of each pixel, which is better suited to the real situation.
Both reasons help to achieve a better performance. 
\begin{figure*}[t]
\noindent \begin{centering}
\subfloat[Urban. The $1{}^{\text{st}}$ row shows the abundance map in pseudo
color; the $2^{\text{nd}}$ row shows the estimated error. \label{fig:AbundMapsOnUrban}]{\noindent \centering{}\includegraphics[width=1.98\columnwidth]{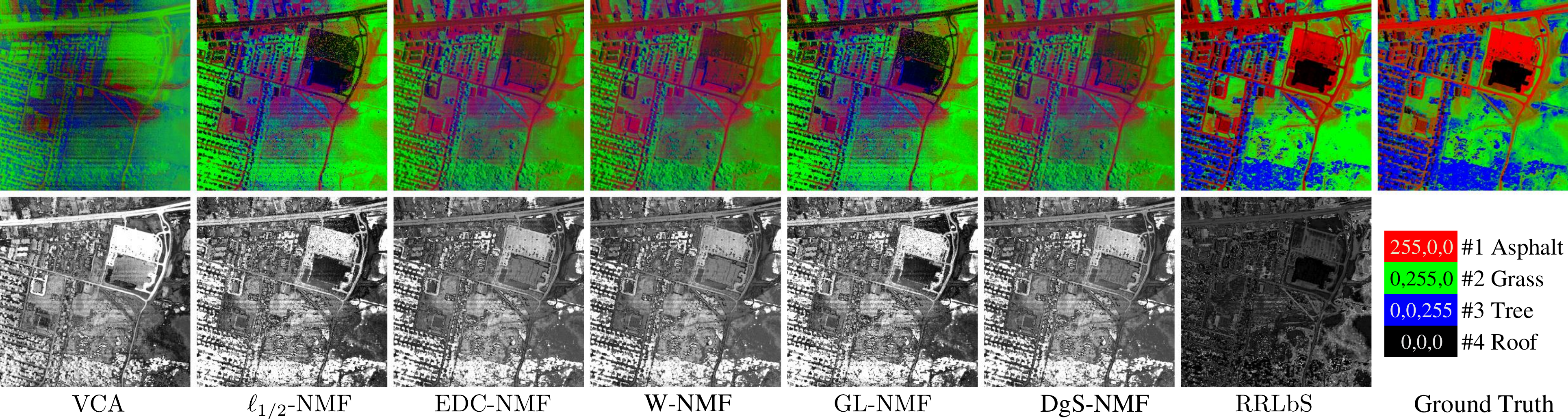}}
\par\end{centering}

\noindent \begin{centering}
\subfloat[Jasper Ridge. The $1{}^{\text{st}}$ row shows the abundance map in
pseudo color; the $2^{\text{nd}}$ row shows the estimated error.
\label{fig:AbundMapsOnJasper}]{\noindent \centering{}\includegraphics[width=1.98\columnwidth]{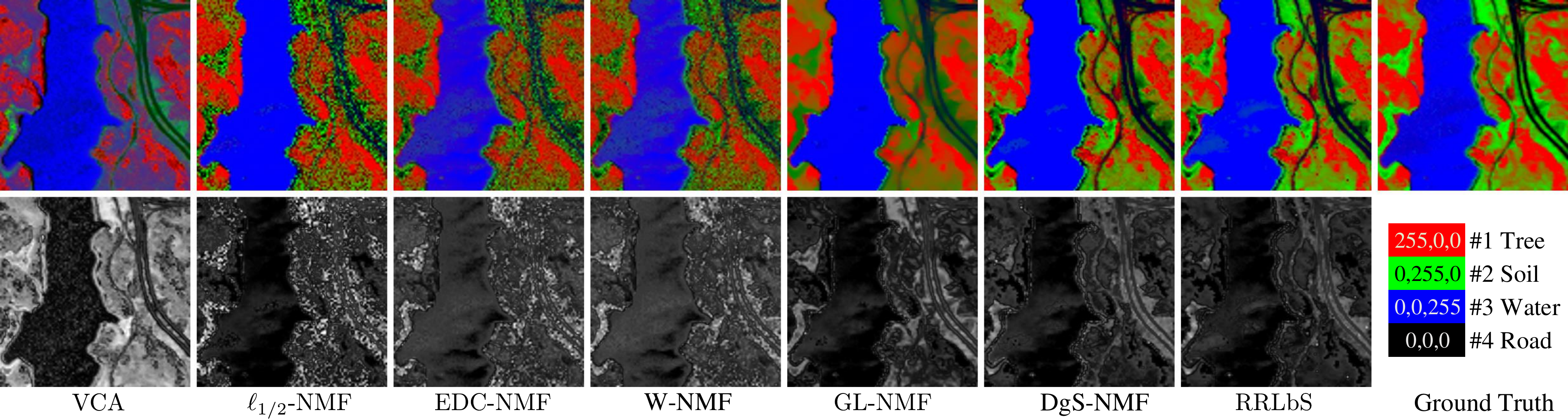}}
\par\end{centering}

\caption{\emph{Abundance} maps of five methods on (a) Urban and (b) Jasper
Ridge. In each sub-figure, the top row shows \emph{abundance} maps
in pseudo color; the bottom row shows the estimated error of each
method, i.e. $\protect\mtbfe=\left\{ e_{n}\right\} _{n=1}^{N}$, where
$e_{n}=\left\Vert \protect\mtbfa_{n}-\protect\mhbfa_{n}\right\Vert $.
From the $1^{\text{st}}$ to the $5^{\protect\mtth}$ column, each
column illustrates the results of one method. The last column shows
the ground truths. (Best viewed in color) \label{fig:AbundMapsOn_UrbanJasperSamson}}
\end{figure*}

In Table~\ref{tab:jasper_SAD=000026RMSE}, there are two sub-tables
illustrating SADs and RMSEs of seven state-of-the-art methods on the
Jasper Ridge hyperspectral image.  In the sub-table, each row shows
the results of one \emph{endmembers}, that is  ``\#1 Tree'', ``\#2
Soil'', ``\#3 Water'' and ``\#4 Road'' respectively. The last row
shows the average results. Specifically, the values in the red ink
are the best, while the blue ones are the second best. As Table~\ref{tab:jasper_SAD=000026RMSE}
shows, our method generally achieves the best results, and in a few
cases it achieves comparable results with the best results of other
methods. Such case is better illustrated in Fig.~\ref{fig:AveragedPerformanceOn3DataSets}b.
As we shall see, RRLbS is the best method that reduces $5.32\%$ for
$\overline{\text{SAD}}$ and $15.45\%$ for $\overline{\text{RMSE}}$
according to the second best results. However, compared with the results
on Urban, the improvement of our method is not so huge. This is since
Jasper Ridge is not so badly noised as Urban. The improvement mainly
relies on the individually sparse constraint in RRLbS. 

Table~\ref{tab:Cuprite_SAD} summaries the HU performances of six
methods on the Cuprite hyperspectral image. In this table, the rows
display the results of 12 \emph{endmembers}, as shown in Table~\ref{tab:Cuprite_SAD}.
In general, the sparsity constrained methods, such as $\ell_{1/2}$-NMF,
GL-NMF and DgS-NMF, obtain better results than the other methods.
This is since sparsity constraints tends to achieve expressive \emph{endmembers}~\cite{Stanzli_01_CVPR_locNMF}.
Such property is more reliable for the HU task. In Fig.~\ref{fig:AveragedPerformanceOn3DataSets}c,
the average performances of $\overline{\text{SAD}}$ are exhibited.
As we shall see, our method obtains superior performances---compared
with the second best results, RRLbS reduces $4.87\%$ for $\overline{\text{SAD}}$.
Cuprite is the most challeging real hyperspectral images. Such improvement
is considerable.

\subsection{Visual Performance Comparisons}

To give a visible comparison, the \emph{abundance} map of seven methods
as well as their estimated errors are compared in Fig.\,\ref{fig:AbundMapsOn_UrbanJasperSamson}.
To begin, we give the definition of \emph{abundance} maps in pseudo
 by taking Fig.\,\ref{fig:AbundMapsOnUrban} as an example. There
are four main color inks in the top row of Fig.\,\ref{fig:AbundMapsOnUrban}.
Via these colors, we could display the \emph{abundances} $A_{kn}$
associated with pixel $\mtbfx_{n}$ by plotting the corresponding
pixel using the proportions of red, blue, green and black inks given
by $A_{kn}$ for $k\!=\!1,2,3,4$, respectively. So, for instance,
a pixel for which $A_{2n}\!=\!1$ will be colored blue, whereas one
for which $A_{2n}\!=\!A_{1n}\!=\!0.5$ will be colored with equal
proportions of red and blue inks and so will appear purple.  In the
bottom row of Fig.\,\ref{fig:AbundMapsOnUrban}, the error map $\mtbfe=\left\{ e_{n}\right\} _{n=1}^{N}\in\mtbbR_{+}$
is displayed. At the $n^{\mtth}$ pixel, the error value in $\mtbfe$
is obtained by computing the $\ell_{2}$-norm of the corresponding
error vector, that is $e_{n}=\left\Vert \mtbfa_{n}-\mhbfa_{n}\right\Vert _{2}$.

The visualized \emph{abundances} on the Urban hyperspectral image
are illustrated in Fig.\,\ref{fig:AbundMapsOnUrban}, consisting
of two rows and eight columns of sub-images. The top row shows the
\emph{abundance} map in pseudo, and the bottom rows shows the corresponding
error maps. From the $1^{\text{st}}$ to the $7^{\mtth}$ columns,
each column shows the result sub-image of one method. The last column
shows the ground truth. As we shall see, our method achieves extraordinarily
results. It gets the most similar \emph{abundance} maps compared with
the ground truth; our error map is the smallest. For the other methods,
they achieve \emph{abundance} maps that have more errors, which are
clearly demonstrated in the corresponding error map. As mentioned
before, it is the serious noise in Urban makes other results bad.
While for our method, the robust objective could greatly relieve the
side effects of outlier channels. Thus, the performance is largely
enhanced. 

The \emph{abundance }maps of seven state-of-the-art methods on the
Jasper Ridge image are displayed in Figs.\,\ref{fig:AbundMapsOnJasper}.
Compared with Fig.\,\ref{fig:AbundMapsOnUrban}, most methods achieve
acceptable results. This is due to the less noise in the Jasper Ridge
image. Specifically, our method achieves much better results than
the other methods---our \emph{abundance} map is highly similar with
the ground truth; the corresponding error map is very small. Such
results verify that the individually sparse constraint is very reliable,
and that RRLbS is well suited to the HU task.

\subsection{\label{sub:Convergence-Study-1}Comparision of Guidance Maps: Quantitative
\& Visual }

As a significant characteristic, RRLbS learns the guided map to model
the individually mixed level of each pixel. Based on the learnt guided
map, we impose the individually sparse constraint according to the
mixed level of each pixel. These two phases help each other to achieve
better and better results. The unmixing results have already been
compared. In this section, the learnt guided maps are systematically
compared. 

\begin{figure}[t]
\begin{centering}
\subfloat[Quantitative results of the three guided maps on Urban and Jasper
Ridge. \label{fig:guidanceCompare_quantitative}]{\centering{}\includegraphics[width=0.98\columnwidth]{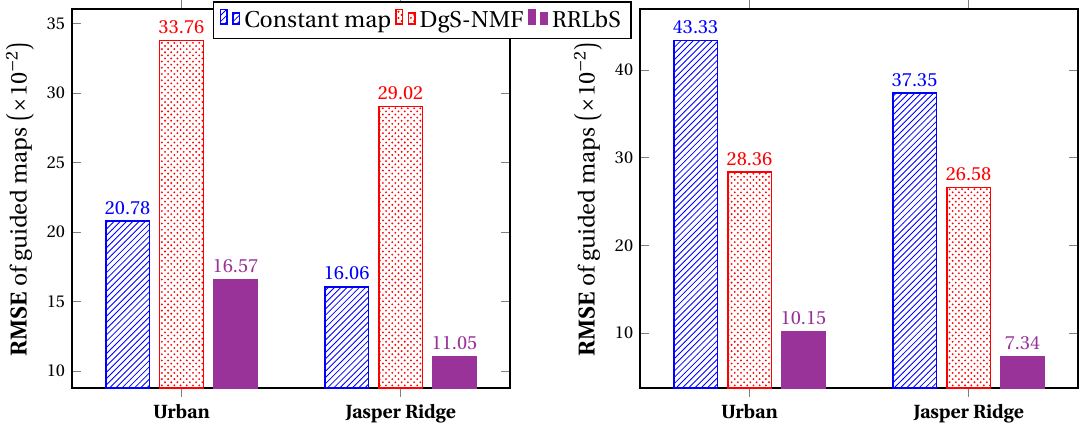}}
\par\end{centering}

\begin{centering}
\subfloat[Visual results of the three guided maps on Urban and Jasper Ridge.\label{fig:guidanceCompare_visual}]{\centering{}\includegraphics[width=0.98\columnwidth]{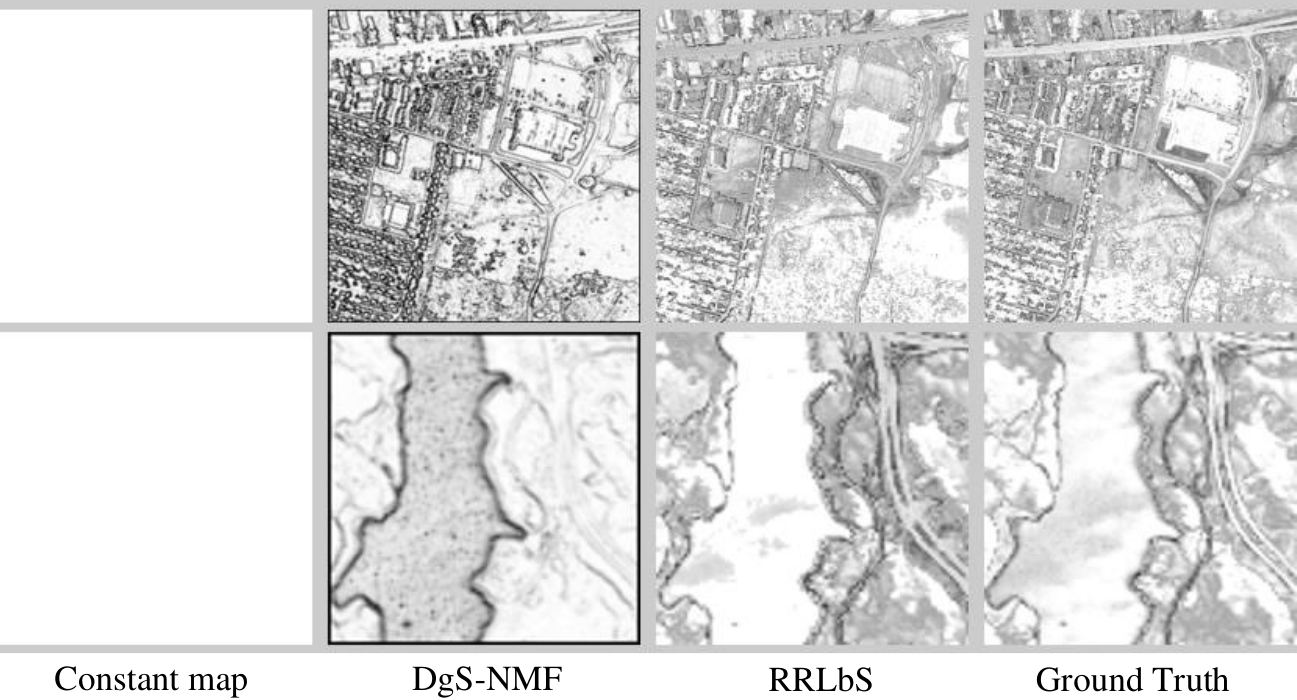}}
\par\end{centering}

\caption{Illustrations of three guided maps on the two datasets: (a) quantitative
results, (b) visual results. Specifically, ``Constant map'' means
the identical guided map used in the traditional methods, e.g., $\ell_{1}$-NMF
and $\ell_{1/2}$-NMF; ``DgS-NMF'' denotes the guided map obtained
by the heuristic strategy in\ \cite{fyzhu_2014_TIP_DgS_NMF}; ``RRLbS''
represents the learning-based guided map achieved by the proposed
method. (a) shows the SAD and RMSE error of those three methods compared
with the ground truths. In (b), there are two rows and four columns
of sub-images. Each row shows the results on one dataset, i.e. Urban
and Jasper Ridge respectively. From the $1^{\text{st}}$ to the $3^{\text{rd}}$
column, each column illustrates the results of one method. The last
column shows the ground truths. \label{fig:guidanceCompare_quantitativeVisual}}
\end{figure}

The results of the guided maps are illustrated in Figs.\ \ref{fig:guidanceCompare_quantitative}
and\ \ref{fig:guidanceCompare_visual}, where the former summarizes
the quantitative results and the latter shows the visual comparisons.
There are three kinds of guided maps in Fig.\ \ref{fig:guidanceCompare_quantitativeVisual}:
1) ``Constant map'' means the identical guided map used by the traditional
methods, like, $\ell_{1}$-NMF and $\ell_{1/2}$-NMF\ \cite{fyzhu_2014_TIP_DgS_NMF};
2) ``DgS-NMF'' denotes the guided learnt by the heuristic strategy
in\ \cite{fyzhu_2014_TIP_DgS_NMF}; 3) ``RRLbS'' represents the learning-based
guided map obtained by our RRLbS. As shown in Fig.\ \ref{fig:guidanceCompare_quantitativeVisual},
``RRLbS'' is extraordinarily better than the other two methods. In
terms of quantitative comparisons (cf. Fig.\ \ref{fig:guidanceCompare_quantitative}),
the estimated error of ``RRLbS'' is half of the second best one in
average. When checking the visual comparison in Figs.\,\ref{fig:guidanceCompare_visual},
our method achieves the most similar appearance compared with the
ground truth. Obviously, there is no meaning for the ``Constant map''.
For the ``DgS-NMF'', it achieves acceptable results in the transitional
areas in the scene. However, it is intractable to guess the mixed
information for the vast smooth areas. As a huge improvement, the
``RRLbS'' gets satisfactory mixed information in all kinds of areas.
In short, the comparisons above verify that RRLbS is able to learn
satisfactory guidance maps which can effectively indicate the mixed
level of pixels. 
\begin{figure}[t]
\begin{centering}
\includegraphics[width=0.99\columnwidth]{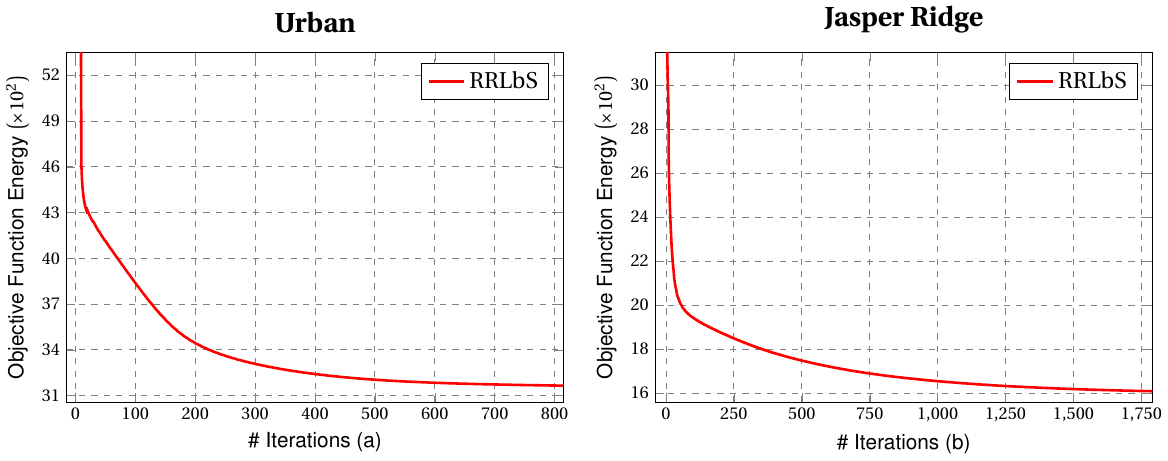}
\par\end{centering}

\caption{Convergence curves of RRLbS on (a) Urban and (b) Jasper Ridge. \label{fig:ConvergenceCurves}}
\end{figure}

\subsection{\label{sub:Convergence-Study}Convergence Study}

It has been theoretically proven that the objective~\eqref{eq:RRLbS_final_model}
is able to converge to a local minimum by using the updating rules~\eqref{eq:RRLbS_update_M}
and~\eqref{eq:RRLbS_update_A}. To verify this conclusion, we conduct
experiments to show the empirical convergence property of RRLbS. The
convergent curves are illustrated in Fig.~\ref{fig:ConvergenceCurves},
including two sub-figures, each of which shows the results on one
dataset. In each sub-figure, the X-axis shows the number of iteration
$t$, and the Y-axis illustrates the objective energy defined in\ \eqref{eq:RRLbS_final_model}.
As we shall see, the objective energy decreases monotonously over
the iteration steps until convergence.

\section{\label{sec:Conclusions}Conclusions}

In this paper, we propose a novel robust representation and learning-based
sparsity (RRLbS) based method for the HU task. The $\ell_{2,1}$-norm
is exploited to measure the representation loss, enhancing the robustness
against outlier channels. Then, through the learning-based sparsity
method, the sparse constraint is adaptively applied according to the
mixed level of each pixel. Such case not only agrees with the practical
situation but also leads the \emph{endmember} toward some spectra
resembling the highly sparse regularized pixel. Extensive experiments
on three benchmark datasets verify the advantages of RRLbS: 1) in
terms of both quantitative and visual performances, RRLbS achieves
extraordinarily better results than all the compared methods; 2) the
estimated guidance map is highly promising as well, providing a more
accurate sparse constraint at the  pixel level. Moreover, both theoretic
proof and empirical results verify the convergence of our method.

\appendix{}

In this appendix, we will provide a $\ell_{2,p}$-norm based robust
model to deal with the badly degraded channel. Specifically in Section\,\ref{sub:L21_robustModel},
we have proposed the $\ell_{2,1}$-norm based measure for the representation
error, leading to the following objective
\begin{align}
\min_{\mtbfM\geq\mtbfzero,\mtbfA\geq\mtbfzero} & \mtcalO=\frac{1}{2}\left\Vert \mtbfX-\mtbfM\mtbfA\right\Vert _{2,1}+\lambda\left\Vert \mtbfA^{\mtbfone-\mtbfH}\right\Vert _{1}.\label{eq:RRLbS_final_model-1}
\end{align}
However, there are theoretical and empirical evidences to demonstrate
the fact that compared with $\ell_{2}$ or $\ell_{1}$ norms, the
$\ell_{\!p}\left(0<p<1\right)$-norm is more able to prevent outliers
from dominating the objective, enhancing the robustness\ \cite{fpNie_2012_ICDM_robustMatrixCompletion}.
Therefore, we provide another new model by using the $\ell_{2,p}\left(0<p<1\right)$-norm
to measure the representation loss
\begin{align}
\min_{\mtbfM\geq\mtbfzero,\mtbfA\geq\mtbfzero} & \mtcalO=\frac{1}{2}\left\Vert \mtbfX-\mtbfM\mtbfA\right\Vert _{2,p}+\lambda\left\Vert \mtbfA^{\mtbfone-\mtbfH}\right\Vert _{1},\label{eq:RRLbS_L2p_model}
\end{align}
where $\left\Vert \mtbfX-\mtbfM\mtbfA\right\Vert _{2,p}\!=\!\sum_{l}^{L}\left(\sum_{n}^{N}\left(\mtbfX-\mtbfM\mtbfA\right)_{ln}^{2}\right)^{p/2}$.
In this case, we could control the robustness level of our model\ \eqref{eq:RRLbS_L2p_model}
by setting the value of $p$---a smaller $p$ leads to a strong robustness
 under the same $\lambda$ setting. 

{\footnotesize{}\bibliographystyle{IEEEtran}
\bibliography{11_home_fyzhu_link2dropbox_self_Folder_myWorksOnDropboxs_bibFiles_referenceBib,12_home_fyzhu_link2dropbox_self_Folder_myWorksOnDropboxs_bibFiles_referenceBib2}
}
\end{document}